%% file: main.tex
\documentclass[10pt]{article} 
\usepackage[preprint]{tmlr}

\input{macros}

\usepackage[backref]{hyperref}
\usepackage{url}
\usepackage{comment}
\usepackage[ruled]{algorithm2e}
\usepackage{amsthm}
\usepackage{graphicx}
\usepackage{subcaption}
\usepackage{adjustbox}
\usepackage{multirow}
\usepackage{enumitem}
\usepackage{longtable}

\newtheorem{definition}{Definition}
\newtheorem{theorem}{Theorem}
\newtheorem*{theorem*}{Theorem}
\newtheorem{lemma}{Lemma}
\newtheorem*{lemma*}{Lemma}

\title{Node-Level Differentially Private Graph Neural Networks}

\def\month{MM}  
\def\year{YYYY} 
\def\openreview{\url{https://openreview.net/forum?id=XXXX}} 

\author{\name Ameya Daigavane \email
ameya.d.98@gmail.com \\
\addr Google Research
\AND
\name Gagan Madan \email
gaganmadan@google.com \\
\addr Google Research
\AND
\name Aditya Sinha \email
adityaasinha28@gmail.com \\
\addr Google Research
\AND
\name Abhradeep Guha Thakurta \email
athakurta@google.com \\
\addr Google Research
\AND
\name Gaurav Aggarwal \email
gauravaggarwal@google.com \\
\addr Google Research
\AND
\name Prateek Jain \email
prajain@google.com \\
\addr Google Research
}

\begin{document}

\maketitle

\input{abstract}

\input{introduction}

\input{related}

\input{preliminaries}

\input{sampling}

\input{method}
\input{exps}

\input{future-work}

\bibliography{reference}
\bibliographystyle{tmlr}

\newpage

\appendix
\input{appendix}
\end{document}

%% file: macros.tex
\usepackage{amsmath,amsfonts,bm}

\newcommand{\newterm}[1]{{\bf #1}}

\def\eqref#1{equation~\ref{#1}}

\def\1{\bm{1}}

\DeclareMathAlphabet{\mathsfit}{\encodingdefault}{\sfdefault}{m}{sl}
\SetMathAlphabet{\mathsfit}{bold}{\encodingdefault}{\sfdefault}{bx}{n}

\def\gA{{\mathcal{A}}}
\def\gB{{\mathcal{B}}}

\newcommand{\E}{\mathbb{E}}

\newcommand{\R}{\mathbb{R}}

\DeclareMathOperator*{\argmin}{arg\,min}

\newcommand{\bfA}{\ensuremath{\mathbf{A}}}

\newcommand{\bfE}{\ensuremath{\mathbf{E}}}

\newcommand{\bfX}{\ensuremath{\mathbf{X}}}
\newcommand{\bfY}{\ensuremath{\mathbf{Y}}}

\newcommand{\bfu}{\ensuremath{\mathbf{u}}}
\newcommand{\bfv}{\ensuremath{\mathbf{v}}}

\newcommand{\bfy}{\ensuremath{\mathbf{y}}}

\newcommand{\bfTheta}{\ensuremath{\mathbf{\Theta}}}

\newcommand{\calA}{\ensuremath{\mathcal{A}}}
\newcommand{\calB}{\ensuremath{\mathcal{B}}}

\newcommand{\calL}{\ensuremath{\mathcal{L}}}

\newcommand{\calN}{\ensuremath{\mathcal{N}}}

\newcommand{\calS}{\ensuremath{\mathcal{S}}}

\newcommand{\hbfv}{\widehat{\bfv}}
\newcommand{\hbfy}{\widehat{\bfy}}
\newcommand{\neighbor}[1]{\calN_{#1}}

\newcommand{\normE}[1]{\left\lVert#1\right\rVert_2}
\newcommand{\normF}[1]{\left\lVert#1\right\rVert_F}

\newcommand{\indeg}{\textup{in-deg}}

\newcommand{\Rd}[3]{D_{#1}(#2 \ \Vert \ #3)}
\newcommand{\ClipC}[1]{\textup{Clip}_C(#1)}
\newcommand{\dT}{\nabla_{\bfTheta}}
\newcommand{\dW}{\dT}

\newcommand{\gnnp}{\mathsf{GNN}(\bfA,\bfX,v; \bfTheta)}

\newcommand{\gnnpt}{\mathsf{GNN}(\bfA,\bfX,v; \bfTheta_t)}

\newcommand{\nm}{\lambda}
\newcommand{\lr}{lr}

\renewcommand{\epsilon}{\varepsilon}

%% file: abstract.tex
\begin{abstract}
Graph Neural Networks (GNNs) are a popular technique for 
modelling graph-structured data and computing node-level representations
via aggregation of information
from the neighborhood of each node.  
However, this aggregation implies increased risk
of revealing sensitive information, 
as a node can participate in the
inference for multiple nodes.
This implies that standard privacy preserving machine learning 
techniques, such as differentially private stochastic gradient descent
(DP-SGD) -- which are designed for situations where each data point
participates in the inference for one point only --
either do not apply, or lead
to inaccurate models. In this work, we formally define the problem of
learning GNN parameters with node-level privacy, and
provide an algorithmic solution with a strong differential privacy
guarantee. We employ a
careful sensitivity analysis and provide a
non-trivial extension of the privacy-by-amplification technique to the GNN setting. 
An empirical evaluation on standard benchmark datasets demonstrates that
our method is indeed able to learn accurate privacy--preserving GNNs which outperform both private and non-private methods that completely
ignore graph information.   
\end{abstract}

%% file: introduction.tex
\section{Introduction}
\label{sec:intro}

\begin{figure}[htbp]
	\centering
    \includegraphics[width=.8\columnwidth]{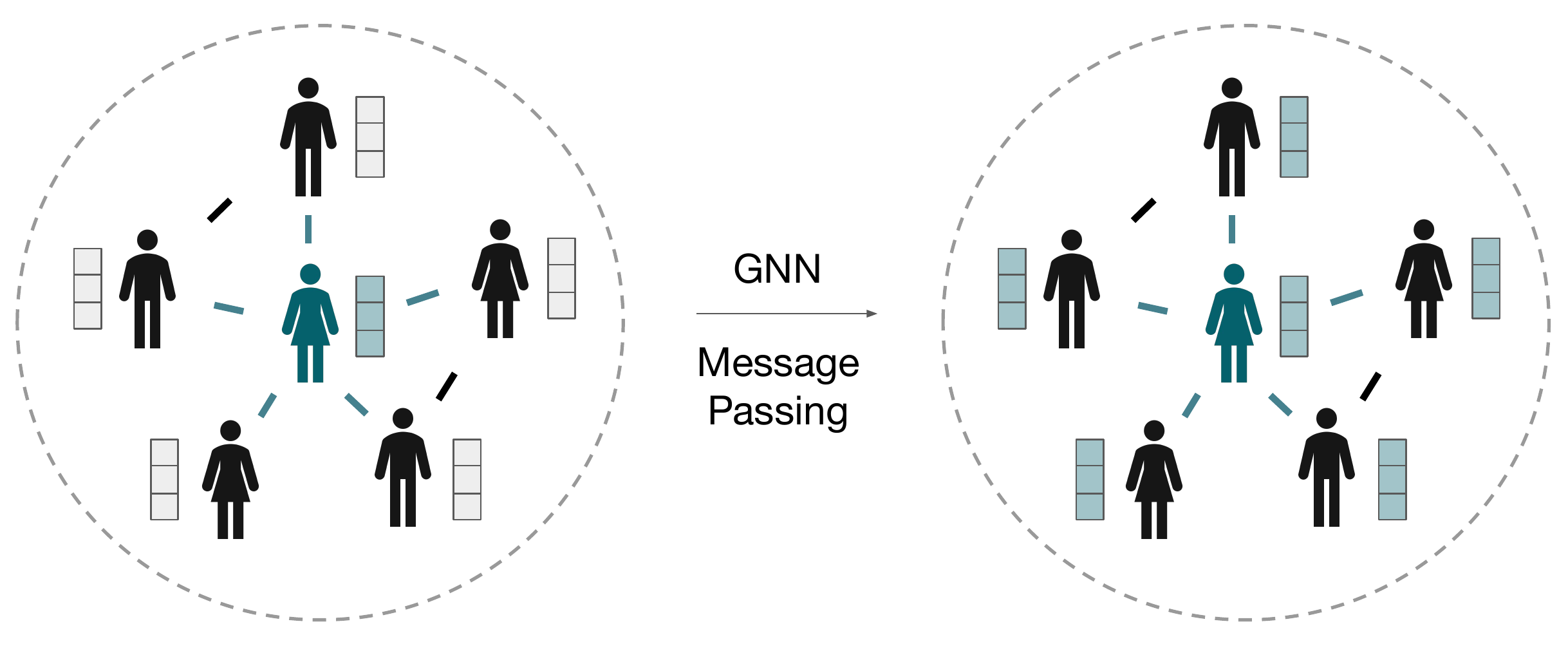}
    \caption{\textbf{In a GNN, every node participates in the predictions for neighbouring nodes, introducing new avenues for privacy leakage. The user corresponding to the center node and all of their private information is highlighted in blue.}
    }\label{fig:cover_pic}
\end{figure}

Graph Neural Networks (GNNs) \citep{kipf:gcn,petar:gat,hamilton:graphsage,gilmer:mpnn}
are powerful modeling tools that capture structural information provided by a graph. Consequently,
they have become popular in a wide array of domains such as
the computational sciences \citep{ktena2018metric,ahmedt2021graph,mccloskey2019using},
computer vision \citep{wang2019dynamic},
and natural language processing \citep{yao2019graph}. GNNs have become an attractive solution for modeling users interacting with each other; each user corresponds to a node of the graph and the user-level interactions correspond to edges in the graph. Thus, GNNs are popular for solving a variety of recommendation and ranking tasks, where it is challenging to obtain and store user data  
\citep{fan2019graph,amar:medres,levy:user-level}. However, such GNN-based solutions are challenging to deploy as they are susceptible to leaking highly sensitive private  information of users. Standard ML models -- without GNN-style neighborhood data aggregation -- are already known to be highly susceptible to leakage of sensitive information about the training data \citep{carlini:memorization}. The risk of leakage of private information is even higher in GNNs as each prediction is based on the node itself and aggregated data from its neighborhood. As depicted in \autoref{fig:cover_pic}, there are two types of highly-sensitive information about an individual node that can be leaked in the GNN setting:
\begin{itemize}
    \item the features associated with the node,
    \item the labels associated with the node, and,
    \item the connectivity (relational) information of the node in the graph.
\end{itemize}

In this work, we study the problem of designing algorithms to learn GNNs while preserving  {\em node}-level privacy.

We use differential privacy as the notion of privacy \citep{dwork:dp} of a node, which requires that the algorithm should learn roughly similar GNN parameters despite
the replacement of an entire node and {\em all} the data points associated with that node.
Our proposed method preserves the privacy of the features of each node, their labels as well as their connectivity information. Our method adapts the standard DP-SGD method \citep{song2013stochastic,BST14,abadi:dp-sgd} to the node-level privacy setting.

DP-SGD is an extension of standard SGD (stochastic gradient descent) that bounds
per-user contributions to the batched gradient
by clipping {\em per-example gradient} terms.
However, standard analysis of DP-SGD does not directly extend to GNNs, as each per-example gradient term in GNNs can depend on private data from {\em multiple nodes}. The key technical contribution of our work is two-fold: 
\begin{itemize}
    \item We propose a graph neighborhood sampling scheme that enables a careful sensitivity analysis for DP-SGD in multi-layer GNNs.
    \item We extend the standard privacy by amplification technique for DP-SGD in multi-layer GNNs, where one per-example gradient term can depend on multiple users.
\end{itemize}

Together, this allows us to learn the
parameters of a GNN with strong {\em node}-level privacy guarantees, as evidenced by empirical results on benchmark datasets in \autoref{sec:experiments}.

%% file: related.tex
\section{Related Work}



Differentially Private SGD (DP-SGD) \citep{song2013stochastic,BST14,abadi:dp-sgd}
has been used successfully to train neural network
models to classify images \citep{abadi:dp-sgd} and text
\citep{anil:dp-bert}, by augmenting 
the standard paradigm of gradient-based training to be
differentially private. Edge-level privacy in GNNs ensures that the existence of an edge between user $i$ and user $j$ does not impact the output significantly \citep{wu:linkteller}. However, such methods do not protect the entirety of user $i$'s private data, listed above. Since each user occurs as a node in the underlying graph, we want to preserve the privacy of {\em all} of a node's data. Private GNNs have also been studied from the
perspective of local privacy \citep{sajadmanesh:lpgnn},
where each node performs its share of the GNN computation locally and sends noisy versions of its data to neighbouring nodes so as to learn shared weights; such  algorithm  needs to correct for the bias in both the features and labels. 
The analysis of this method only applies to GNNs with linear
neighborhood aggregation functions. 
In contrast, the methods we propose can be employed with a large class of GNN models called `message-passing' GNNs, described in \autoref{sec:preliminaries}.
\citep{wu:fedgnn} utilizes private GNNs for recommendation
systems, but their method assumes a bipartite graph structure, and cannot be naturally extended to homogeneous graphs.
Other approaches employ federated learning \citep{zhou:vfgnn}, but only guarantee that the GNN neighbourhood aggregation step is differentially private, which is insufficient to guarantee privacy of each node's neighborhood. 
Finally, several papers provide privacy-preserving GNNs 
\citep{shan:sapgnn} but these do not use the formal notion of DP
and provide significantly weaker privacy guarantees.  In
different contexts, there has been extensive work on node-level
DP ~\citep{raskhodnikova:lipschitz,karwa:graph-structure,borgs:private-graphon,borgs:revealing-graphon}.
But these methods generally estimate `global' graph-level 
statistics and do not support learning methods such as GNNs. 
In contrast, our approach  {\em predicts} `local' node-level statistics (such as the label of a node) while preserving node-level privacy.

%% file: preliminaries.tex
\section{Problem Formulation and Preliminaries}
\label{sec:preliminaries}
Consider a graph dataset $G=(V, E, \bfX, \bfY)$ with {\em directed} graph $\mathcal{G}=(V,E)$ represented by a adjacency matrix $\bfA\in\{0,1\}^{n\times n}$. $n$ is the number of nodes in $\mathcal{G}$, $V$ denotes the node set, $E$ denotes the edge set. Each node $v$ in the graph is equipped with a feature vector $\bfX_v\in \mathbb{R}^d$; $\bfX\in\mathbb{R}^{n\times d}$  denotes the feature matrix. $\bfY\in\mathbb{R}^{n\times Q}$ is the label matrix and $\bfy_v$ is the label for the $v$-th node
over $Q$ classes. Note that many of the labels in the label vector can be missing, which models the semi-supervised setting.
In particular, we assume that node labels $\bfy_v$ are only
provided for a subset of nodes $V_{tr} \subset V$, called the
training set. Given the graph dataset $G$, the goal is to learn parameters of a
GNN while preserving privacy of individual nodes. A
$r$-layer GNN\footnote{This is sometimes referred to as a $r$-hop GNN in the literature.} can be generally represented by the following operations: 
\begin{align}
    \label{eqn:r-layer-gnn}
    \hbfy_v = \gnnp :=
      f_\text{dec}\left(f_\text{agg}
      \left(\{f_{\text{enc}}(\bfX_u) \ | \ \bfA^{(r)}_{vu} \neq 0 \}\right)
      \right)
\end{align}
where $\hbfy_v$ is the prediction from
the GNN for a given node $v$,
$f_\text{enc}$ is the encoder function
that encodes node features
with parameters $\Theta_{\text{enc}}$,
$f_\text{agg}$ is the neighborhood aggregation
function with parameters $\Theta_{\text{agg}}$, $f_\text{dec}$
is the prediction decoder function with parameters
$\Theta_{\text{dec}}$,
and $\Theta:=(\Theta_\text{enc}, \Theta_\text{agg},
\Theta_\text{dec})$.

As seen from \autoref{eqn:r-layer-gnn}, we can represent an $r$-layer GNN by
a convolution over the local $r$-hop neighborhood 
of each node, represented as a subgraph of $G$ rooted at the node.

While our results apply to most classes of GNN models
\citep{hamilton:graphsage,petar:gat,xu:gin}, for simplicity, we focus on
Graph Convolutional Network (GCN) \citep{kipf:gcn}
with additional results for the 
Graph Isomorphism Network (GIN) \citep{xu:gin}
and the Graph Attention Network (GAT) \citep{petar:gat} in the Appendix. Thus, `learning' a GNN is equivalent to finding parameters
$\Theta:=(\Theta_\text{enc}, \Theta_\text{agg},
\Theta_\text{dec})$ that
minimize a suitable loss: 
\begin{align}
    \Theta^{*} = \argmin_{\Theta} \calL(G,\Theta):= \sum\limits_{v\in V} \ell(\hbfy_v; \bfy_v), 
    \label{eq:loss}
\end{align}
where $\ell:\mathbb{R}^{Q\times Q}\rightarrow \mathbb{R}$ is a standard loss function such as categorical or sigmoidal cross-entropy.

\begin{definition}[Adjacent Graph Datasets]
Two graph datasets $G$ and $G'$ are said to be
\newterm{node-level adjacent}
if one can be obtained by adding
or removing a node
(with its features, labels and associated edges)
to the other. That is, $G$ and $G'$ are exactly the same except for the $v$-th node, i.e., $\bfX_v$, $\bfy_v$ and $\bfA_v$ differ in the two datasets. 
\end{definition}
Informally, $\gA$ is said to be node-level differentially-private if the
addition or removal of a node in $\gA$'s input does not affect $\gA$'s
output significantly. 
\begin{definition}[Node-level Differential Privacy]\label{def:rdp}
Consider any randomized algorithm
$\gA$ that takes as input a graph dataset. 
$\gA$ is said to be $(\alpha, \gamma)$
\newterm{node-level R\'enyi differentially-private} \citep{mironov2017renyi}
if, for every pair of node-level adjacent datasets $G$ and $G'$: $\Rd{\alpha}{\gA(G)}{\gA(G')} \leq \gamma$
where \newterm{R\'enyi divergence} $D_\alpha$ of order $\alpha$ between two random variables $P$ and $Q$
is defined as
$\Rd{\alpha}{P}{Q} = \frac{1}{\alpha - 1} \ln \E_{x\sim Q}\left[\frac{P(x)}{Q(x)}\right]^\alpha.$
\end{definition}
\vspace*{-5pt}
Note that we use R\'enyi differentially-private (RDP) \citep{mironov2017renyi} as the formal notion of differential privacy (DP), as it allows for tighter composition of DP across multiple steps. This notion is closely related to the standard $(\epsilon,\delta)$-differential privacy \citep{dwork:dp}; Proposition 3 of \cite{mironov2017renyi} states that any $(\alpha, \gamma)$-RDP
mechanism also satisfies $(\gamma +
\frac{\log 1/\delta}{\alpha - 1}, \delta$)-differential privacy
for any $\delta \in (0, 1)$. Thus, we seek to find $\Theta$ by optimizing \autoref{eq:loss} while
ensuring RDP.
\begin{definition}
    The \newterm{$K$-restricted node-level sensitivity} $\Delta_K(f)$
    of a function $f$ defined on graph datasets is $\Delta_K(f) =
        \max_{\substack{
         G, G' \text{ node-level adjacent} \\
        \indeg(G), \ \indeg(G') \leq K }}
        \normE{f(G) - f(G')}.$
        
    \label{def:nds}
\end{definition}
\vspace*{-5pt}


%% file: sampling.tex
\section{Sampling Subgraphs with Occurrence Constraints}
\label{app:sampling}
To bound the sensitivity of the mini-batch gradient in
\autoref{alg:dp-sgd}, we must carefully bound the 
maximum number of occurrences
of any node in the graph across all training subgraphs.
To ensure that
these constraints are met for any $r$-layer GNN,
we propose
$\mathsf{SAMPLE-SUBGRAPHS}$ (\autoref{alg:sample-subgraphs})
to output a set of training subgraphs.
Note that the common practice \citep{hamilton:graphsage}
of sampling to restrict the out-degree of every node
is insufficient to provide such a guarantee,
as the in-degree of a node
(and hence, the number of occurrences of that node in other subgraphs)
can be very large. Once the model parameters have been learnt,
no restrictions are needed at inference time. This means GNN predictions for the `test' nodes can use the entire neighbourhood information.

\begin{algorithm}[htbp]
\caption{$\mathsf{SAMPLE-EDGELISTS}$:
Sampling the Adjacency Matrix with In-Degree Constraints}
\KwData{Graph $G = (V, E, \bfX, \bfY)$, Training set $V_{tr}$,
Maximum in-degree $K$.}
\KwResult{Set of sampled edgelists $\bfE_v$ for each node $v \in V$.}
\For{$v \in V$}{
Construct the incoming edgelist over training set:
$\mathbf{RE}_v \gets \{u \mid (u, v) \in E \text{ and } u \in V_{tr}\}$ \\
Sample incoming edgelists. Each edge is sampled
independently with a probability $p = \frac{K}{2|\mathbf{RE}_v|}$: 
$\mathbf{RE}_v \gets \text{sample}(\mathbf{RE}_v)$ \\
The nodes with in-degree greater than $K$ are dropped from all
edgelists.\footnotemark \\
}
\For{$v \in V$}{
Reverse incoming edgelists to get sampled edgelists $\mathbf{E}_v$:
$\mathbf{E}_v \gets \{u \mid v \in \mathbf{RE}_u \}$
\\
}
\textbf{return} $\{\bfE_v \mid v \in V\}$.
\label{alg:sample-edgelists}
\end{algorithm}

\begin{algorithm}[htbp]
\caption{$\mathsf{DFS-TREE}$:
Depth-First-Search Tree with Depth Constraints
}
\KwData{Root node $v$, Edgelists $\bfE$, Maximum depth $r$.}
\KwResult{Subgraph $S_{v}$ representing the $r$-depth DFS tree rooted at $v$.}
If $r = 0$, \textbf{return} $\{v\}$.\\
Add edges from $v$ to its neighbours' subgraphs:
$S_v \gets \{v\} \cup \{\mathsf{DFS-TREE}(u, \bfE, r - 1) \ | \ u \in \bfE_v\}$ \\
\textbf{return} $S_v$.
\label{alg:dfs-tree}
\end{algorithm}

\begin{algorithm}[htbp]
\caption{$\mathsf{SAMPLE-SUBGRAPHS}$:
Sampling Local Neighborhoods with Occurrence Constraints}
\KwData{Graph $G = (V, E, \bfX, \bfY)$, Training set $V_{tr}$,
Maximum in-degree $K$, GNN layers $r$.}
\KwResult{Set of subgraphs $S_{v}$ for each node $v \in V_{tr}$.}

Obtain set of sampled edgelists: $\bfE \gets \mathsf{SAMPLE-EDGELIST}(G, V_{tr}, K)$ \\
\For{$v \in V_{tr}$}{
  $S_v \gets \mathsf{DFS-TREE}(v, \bfE, r)$
}
\textbf{return} $\{S_v \mid v \in V_{tr}\}$.
\label{alg:sample-subgraphs}
\end{algorithm}

\begin{lemma}[$\mathsf{SAMPLE-SUBGRAPHS}$ Satisfies Occurrence Constraints]
\label{lm:sampling}
Let $G$ be any graph
with set of training nodes $V_{tr}$.
Then, for any $K, r \geq 0$, the number of occurrences of any node
in the set of training subgraphs
$\mathsf{SAMPLE-SUBGRAPHS}(G, V_{tr}, K, r)$
is bounded above by $N(K, r)$, where:
$$
N(K, r) = \sum_{i = 0}^r K^i = \frac{K^{r + 1} - 1}{K - 1} \in \Theta(K^r)$$
\end{lemma}

In the interest of space, we have supplied the proof of \autoref{lm:sampling} in \autoref{app:sampling-proof}.

%% file: method.tex
\section{Learning Graph Neural Networks (GNNs) via DP-SGD}
\label{sec:main-algorithm}

In this section, we provide a variant of DP-SGD~\citep{BST14} designed specifically for GNNs, and show that our method guarantees node-level DP (Definition~\ref{def:rdp}). Assuming we are running a $r$-layer GNN,
we first subsample the local $r$-hop neighborhood of each
node to ensure that each node has 
a bounded number of neighbors and influences a small number of nodes.
Next, similar to the standard mini-batch SGD technique, we sample a subset $\calB_t$ of $m$ subgraphs chosen uniformly at random from the set $\calS_{tr}$ of training subgraphs. In contrast to the standard mini-batch SGD, that samples points with replacement for constructing a mini-batch, our method samples mini-batch $\calB_t$ uniformly from the set of all training subgraphs. This distinction is important for our privacy amplification result. Once we sample the mini-batch, we apply the standard DP-SGD procedure of computing the gradient over the mini-batch, clipping the gradient and adding noise to it, and then use the noisy gradients for updating the parameters. 

However, DP-SGD requires each update to be differentially private. In standard settings where each gradient term in the mini-batch corresponds to only one point, we only need to add $O(C)$ noise -- where $C$ is the clipping norm of the gradient -- to ensure privacy. However, in the case of GNNs with node-level privacy, perturbing one node/point $\hbfv$ can have impact on the loss terms corresponding to all its neighbors. Thus, to ensure the privacy of each update, we add noise according to the sensitivity of aggregated gradient: 
$
\nabla_{\bfTheta}\calL(\calB_t;\bfTheta_t):=\sum_{S_v \in \gB_t}
    \ClipC{\dT\ell\left(\gnnpt;\bfy_v\right)}
$
with respect to any individual node $\hbfv$, which we bound via careful  subsampling of the input graph. In traditional DP-SGD, a crucial component in getting a better privacy/utility trade-off over just adding noise according to the sensitivity of the minibatch gradient, is privacy amplification by sampling~\citep{KLNRS,BST14}. This says that if an algorithm $\calA$ is $\epsilon$-DP on a data set $D_1$, then on a random subset $D_2\subseteq D_1$ it satisfies roughly $ \frac{|D_2|}{|D_1|}\left(e^\epsilon-1\right)$-DP. Unlike traditional ERM problems, we cannot directly use this result in the context of GNNs. The reason is again that on two adjacent data sets, multiple loss terms corresponding to $\hbfv$ and its $r$-hop neighbors $\neighbor{\hbfv}^{(r)}$ get modified. To complicate things further, the minibatch $\calB_t$ that gets selected may only contain a small random subset of $\neighbor{\hbfv}^{(r)}$. To address these issues, we provide a new privacy amplification theorem (\autoref{thm:gnn-amplified-privacy-guarantee}). To prove the theorem, we adapt \cite[Lemma 25]{FMTT18} -- that shows a weak form of convexity of R\'enyi divergence -- for our specific instance, and provide a tighter bound by exploiting the special structure in our setting along with the above bound on sensitivity. 

\begin{algorithm}[t]
\caption{DP-GNN (SGD): Training Differentially Private Graph Neural Networks with SGD}
\KwData{Graph $G = (V, E, \bfX, \bfY)$, GNN model $\mathsf{GNN}$,
Number of GNN layers $r$, Training set $V_{tr}$, Loss function $\calL$, Batch size $m$, Maximum in-degree $K$, Learning rate $\eta$, Clipping threshold $C$, Noise standard deviation $\sigma$, Maximum training iterations $T$.}
\KwResult{GNN parameters $\bfTheta_{T}$.}
 Note that $V_{tr}$ is the subset of nodes for which labels are
 available (see Paragraph 1 of \autoref{sec:preliminaries}). \\
 Construct the set of training subgraphs with \autoref{alg:sample-subgraphs}: $\calS_{tr} \gets \mathsf{SAMPLE-SUBGRAPHS}(G, V_{tr}, K, r)$.\\
 Initialize $\bfTheta_0$ randomly. \\
 \For{$t = 0$ \KwTo $T$}{
  Sample set $\calB_t \subseteq \calS_{tr}$ of size $m$ uniformly at random from all subsets of $\calS_{tr}$.\\
  Compute the update term $\bfu_t$ as the sum of the clipped gradient terms in the mini-batch $\gB_t$:
  $
    \bfu_t \gets \sum_{S_v \in \gB_t}
    \ClipC{\dT\ell\left(\gnnpt;\bfy_v\right)}
  $ \\
  Add independent Gaussian noise to the update term:
  $\Tilde{\bfu}_t \gets \bfu_t + \calN(0, \sigma^2 \mathbb{I}) $ \\
  Update the current estimate of the parameters with the noisy update: 
  $\bfTheta_{t + 1} \gets \bfTheta_{t} - \frac{\eta}{m} \Tilde{\bfu}_t$
 }
 \label{alg:dp-sgd}
\end{algorithm}



\begin{theorem}[Amplified Privacy Guarantee for any $r$-Layer GNN]
\label{thm:gnn-amplified-privacy-guarantee}
Consider the loss function $\calL$ of the form: 
$$
\calL(G, \bfTheta) = \sum_{v \in V_{tr}} \ell\left(\gnnpt;\bfy_v\right).
$$
Recall, $N$ is the number of training nodes $V_{tr}$, $K$ is the maximum in-degree of the
input graph, $r$ is the number of GNN layers,
and $m$ is the batch size.
For any choice of the noise standard
deviation $\sigma > 0$ and clipping threshold $C$, every iteration $t$ of \autoref{alg:dp-sgd}
is $(\alpha, \gamma)$ node-level
R\'enyi DP, where:
\begin{align*}
\begin{split}
\gamma = \frac{1}{\alpha - 1} \ln \E_{\rho}\left[\exp{\left(\alpha (\alpha - 1) \cdot \frac{2\rho^2C^2}{\sigma^2}\right)}\right], \ \rho \sim \textup{Hypergeometric}\left(N, \frac{K^{r + 1} - 1}{K - 1}, m\right).
\end{split}
\end{align*}
$\textup{Hypergeometric}$ denotes the standard hypergeometric distribution \citep{forbes:statistical}. By the standard composition theorem for R\'enyi Differential Privacy~\citep{mironov2017renyi}, over $T$ iterations, \autoref{alg:dp-sgd} is $(\alpha, \gamma T)$ node-level R\'enyi DP, where $\gamma$ and $\alpha$ are defined above.
\end{theorem}

In the interest of space, we have supplied the proof of \autoref{thm:gnn-amplified-privacy-guarantee} in \autoref{app:gnn-amplified-privacy-guarantee-proof}.


{\bf Remark 1}: 
Roughly, for a 1-layer GNN with $m \gg K$, the above bound implies $\sigma=O(mK/N)$ noise to be added per update step to ensure R\'enyi DP with $\alpha=O(1)$ and $\gamma=O(1)$. Note that the standard DP-SGD style privacy amplification results do not apply to our setting as each gradient term can be impacted by multiple nodes. 

{\bf Remark 2}: We provide node-level privacy, which means that our method preserves the neighborhood information of every node. But, we require a directed graph structure, so that changing a row in the adjacency matrix does not impact any other part of the matrix. This is a natural assumption in a variety of settings. For example, when the graph is constructed by `viewership' data in social networks, following the rule that edge $(v, v')$ exists iff user $v$ viewed a post from user $v'$.

{\bf Remark 3}: The careful reader may notice that the sampling procedure ensures that the number of gradient terms affected by the removal of a single node is bounded. A natural question to ask is then, can we directly use group privacy guarantees provided by RDP for our analysis? The answer is yes; however, the resulting bounds are much weaker, because the privacy guarantee of group privacy scales exponentially with the size of the group \citep{mironov:rdp}, which is $N(K, r)$ here. In comparison, \autoref{thm:gnn-amplified-privacy-guarantee}
guarantees that the privacy guarantee scales  only linearly with $N(K, r)$.

{\bf Remark 4}: For a similar reason as above, and since privacy amplification by subsampling requires uniform sampling over all nodes (and their corresponding subgraphs)
and not edges, one cannot use group privacy guarantees with edge-level differentially private GNNs to obtain an equivalent of \autoref{thm:gnn-amplified-privacy-guarantee}.

We similarly adapt a DP version of the Adam \citep{kingma2014adam} optimizer to the GNN setting, called DP-GNN (Adam), with the same privacy guarantees as DP-GNN (SGD).

%% file: exps.tex
\section{Experimental Results}
\label{sec:experiments}
In this section, we present empirical evaluation of our method on standard benchmark datasets for large graphs from the Open Graph Benchmark (OGB) suite \citep{hu2020open} and GraphSAGE \citep{hamilton:graphsage}, and evaluate our method in both transductive and inductive settings. The goal is to demonstrate that our method (DP-GNN) can indeed learn privacy preserving GNNs accurately. 
In particular, we benchmark the following methods:
\begin{itemize}
    \item \textbf{DP-GCN}: Our DP-GNN method (\autoref{alg:dp-sgd}) applied to a $1$-layer GCN (in the transductive and inductive settings) and a $2$-layer GCN (in the inductive settings) with an MLP as the encoder and the decoder.
    \item \textbf{GCN}: A $1$-layer GCN (in transductive and inductive settings) and a $2$-layer GCN (in inductive settings) with MLP as the encoder and decoder.
    In general, this non-private GCN model bounds the maximum accuracy we can hope to achieve from our DP-GCN model.
    \item \textbf{MLP}: A standard multi-layer perceptron (MLP) architecture on the raw node features as proposed in prior works  \citep{hu2020open}, which does not utilize any graph information.
    \item \textbf{DP-MLP}: A DP version of a standard MLP trained using DP-Adam.
\end{itemize}

Currently, practitioners can not use
sensitive graph information in data-critical scenarios,
and have to completely discard
GNN-based models due to {\em privacy concerns}.
Below, we demonstrate that DP-GNN is
able to provide {\em more accurate solutions} than standard
methods that completely discard the graph information, 
while guaranteeing {\em node}-level privacy.

\begin{figure}[htbp]
\begin{tabular}{cccc}
     \hspace*{-5pt}\includegraphics[width=.24\textwidth]{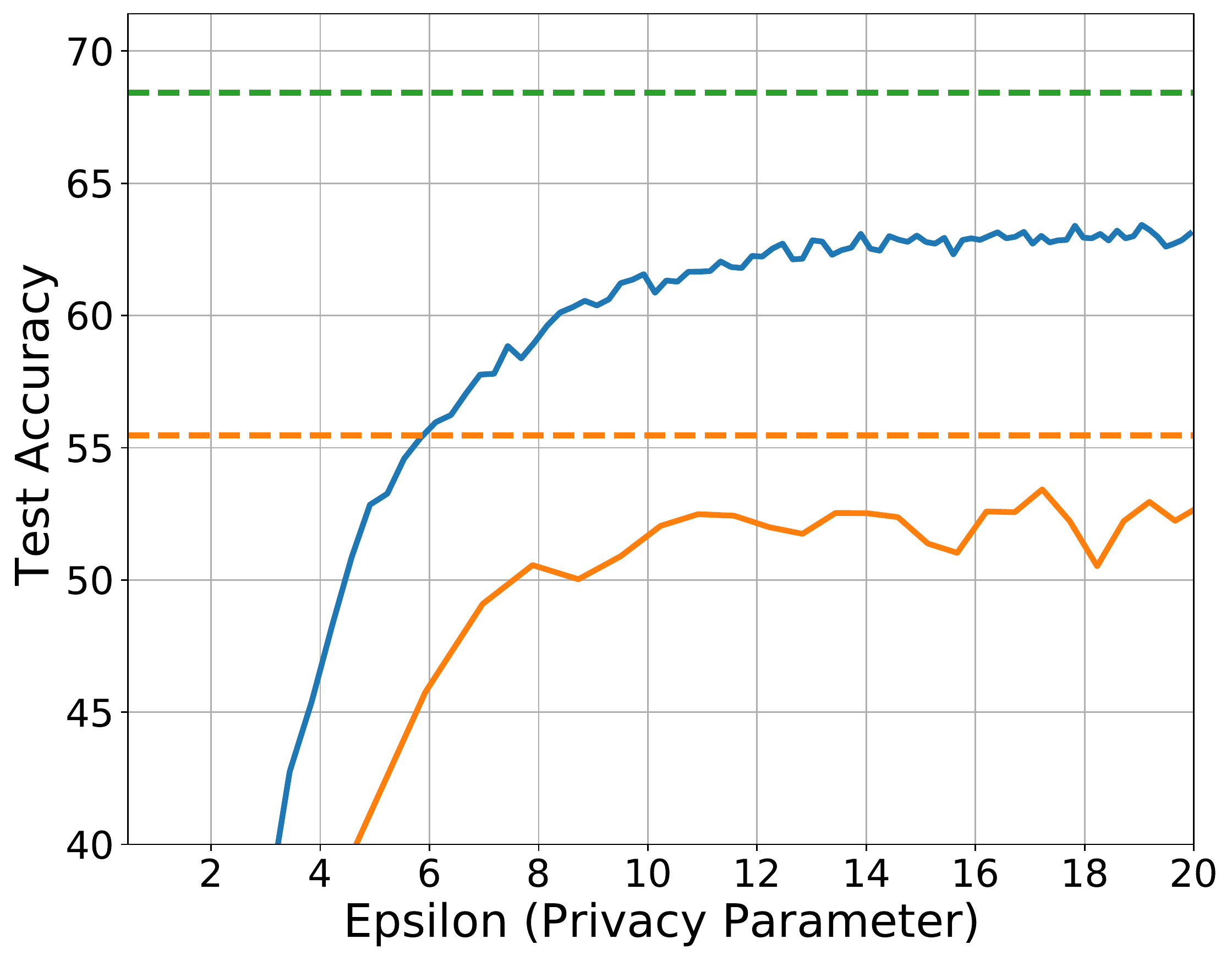} &\hspace*{-5pt} \includegraphics[width=.24\textwidth]{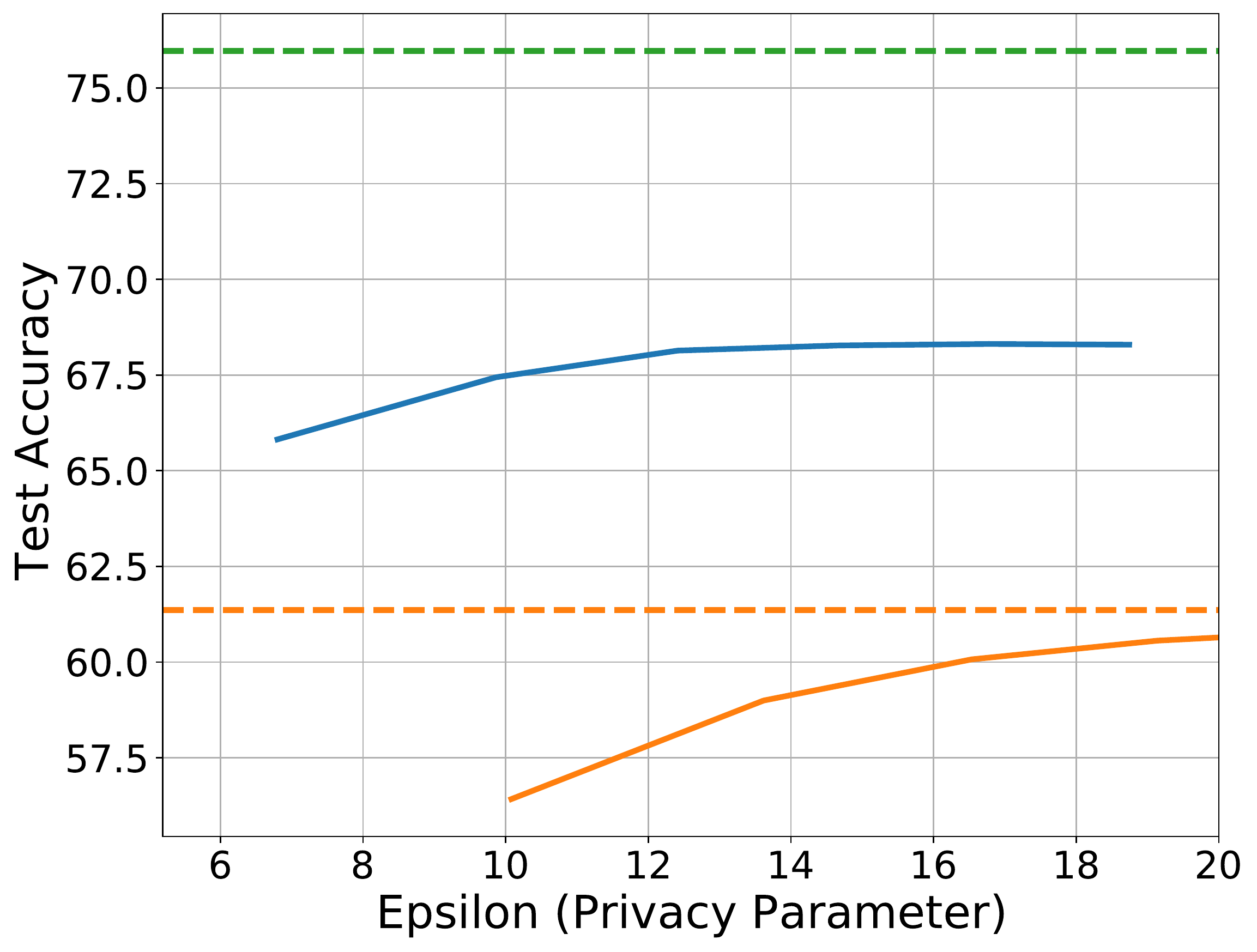} &\hspace*{-5pt} \includegraphics[width=.24\textwidth]{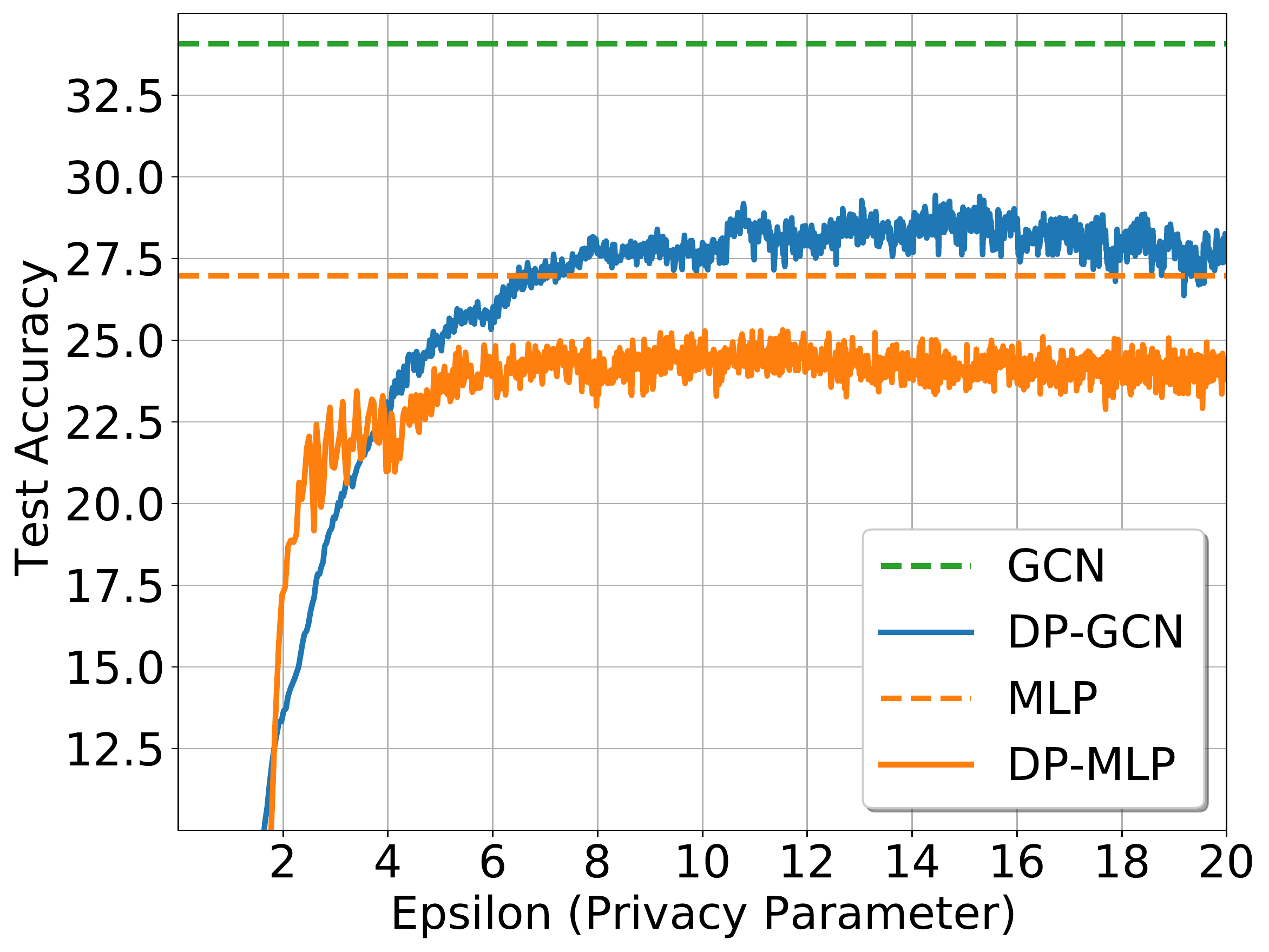} &\hspace*{-5pt} \includegraphics[width=.24\textwidth]{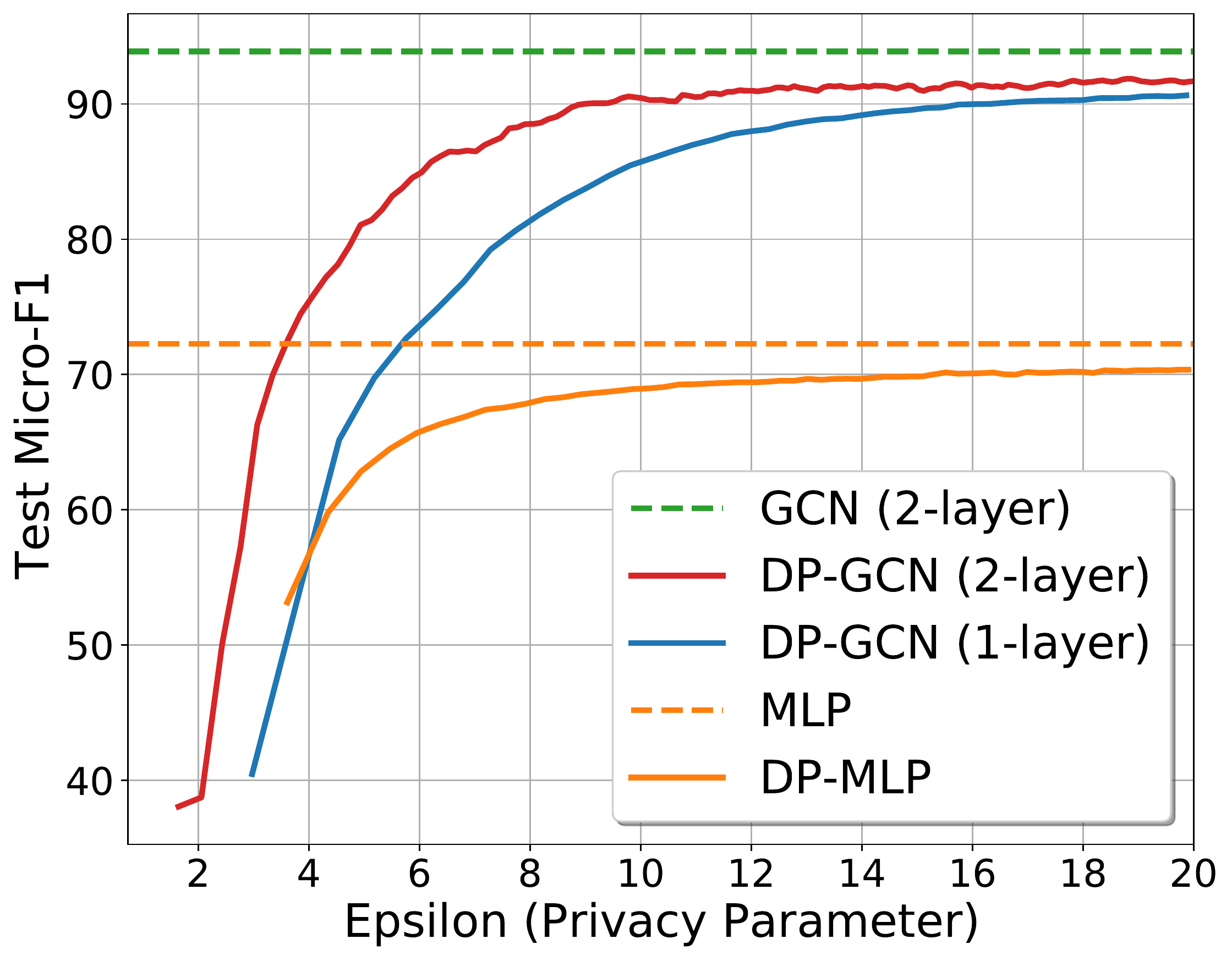}\\[-1pt]
     {\vspace*{-5pt}\bf (a)}&{\bf (b)}&{\bf (c)}&{\bf (d)}\vspace*{-5pt}
\end{tabular}    
\caption{(a), (b), (c): Performance of the 1-layer DP-GCN models and baselines with respect to privacy budget $\epsilon$ on  ogbn-arxiv, ogbn-products and ogbn-mag datasets. (d): Performance of the 1-layer and 2-layer DP-GCN models on the reddit-disjoint dataset.}
    \label{fig:gcn_epsilon}
\end{figure}

\textbf{Gradient Clipping:} For DP-GNN and DP-MLP,  we perform layer-wise gradient clipping: the gradients corresponding to the encoder, aggregation and decoder functions are clipped independently with different clipping thresholds. For each layer, the clipping threshold $C$ in \autoref{alg:dp-sgd} is chosen as the $75\text{th}$ percentile of gradient norms for that layer at initialization on the training data.
We set the noise for each layer $\sigma$ such that the noise multiplier $\nm = \frac{\sigma}{2C \cdot N(K, r)}$ is
identical for each layer, where $\sigma/\nm$ is essentially the sensitivity. It is not hard to observe that the overall privacy cost only depends on $\nm$.

\subsection{Results in the Transductive Setting}

\label{subsec:transductive_results}
\begin{table}[!htbp]
\caption{\textbf{Test performance of DP-GCN in the transductive setting, with privacy budget $\epsilon \leq 12$.}}
\centering
\resizebox{0.85\linewidth}{!}{\begin{tabular}{ccccc}
\hline
Model & ogbn-arxiv & ogbn-products & ogbn-mag & reddit\\
\hline \hline
 GCN (1-layer)                  & 67.758 $\pm$ 0.418 &   75.965 $\pm$ 0.374 &    34.074 $\pm$ 0.445 &	94.074 $\pm$ 0.074 \\
\hline
\textbf{DP}-\textbf{GCN} (Adam) & 61.750 $\pm$ 0.282 &   67.299 $\pm$ 0.103 &    28.933 $\pm$ 0.178 &    92.377 $\pm$ 0.021 \\
\textbf{DP}-\textbf{GCN} (SGD)  & 62.862 $\pm$ 0.977 &   66.954 $\pm$ 0.082 &    28.521 $\pm$ 0.703 &    91.462 $\pm$ 0.098 \\
\hline 
MLP                             & 55.236 $\pm$ 0.317 &   61.364 $\pm$ 0.132 &    26.969 $\pm$ 0.361 &    72.359 $\pm$ 0.141\\
DP-MLP                          & 52.178 $\pm$ 0.158 &   56.449 $\pm$ 0.105 &    25.133 $\pm$ 0.149 &    69.499 $\pm$ 0.104 \\
\hline 
\end{tabular}}
\label{tab:transductive_results}
\end{table} 

We first study the `transductive' setting where the test nodes
are visible during training.
At inference time, each node has access to the features of its
neighbors. Recall that we focus on 1-layer GNNs in this
setting. \autoref{tab:transductive_results} compares the
performance of DP-GCN against baselines on the ogbn-arxiv,
ogbn-products, ogbn-mag and reddit-transductive datasets.
Overall, we observe that our proposed method DP-GCN
significantly outperforms the non-private MLP (which does not use any graph information) and private DP-MLP (which does not use any graph information but trained using standard
DP-Adam) baselines on all of the datasets and with a privacy
budget of $\epsilon\leq 12$. For example,
for ogbn-arxiv and ogbn-products, our method DP-GCN (SGD) is about 6\%
more accurate than MLP and 10\% more accurate than DP-MLP.
For reddit, our method is about 20\% more accurate than MLP
and DP-MLP.
\autoref{fig:gcn_epsilon}
provides a comparison of epsilon (privacy guarantee) versus
test set performance for the three benchmark datasets.
As the privacy budget increases in \autoref{fig:gcn_epsilon},
the performance gap between DP-GCN and the baseline MLP and
DP-MLP widens.
On all datasets, DP-GCN (Adam) outperforms both MLP and DP-MLP
for a privacy budget of $\epsilon \leq 12$.

\subsection{Results in the Inductive Setting}
\label{subsec:inductive_results}

Now, we consider the more challenging `inductive' setting
where the test dataset (the nodes and the graph) are completely disjoint from the training data nodes and the associated graph. This models `multi-enterprise' situations where the graph over users of one enterprise is completely disjoint from the graph over the users of another enterprise.
To conduct these experiments, we divide the nodes into three splits
-- training, validation and test -- and remove all inter-split edges
to partition the graph into disjoint subgraphs.  We report results on three datasets: ogbn-arxiv-disjoint
(where inter-split edges in ogbn-arxiv have been removed),
ogbn-arxiv-clustered
(where agglomerative clustering\footnote{See \autoref{app:reproducibility} for dataset details.} is perfomed on the original ogbn-arxiv dataset to partition the nodes) and
reddit-disjoint (where inter-split edges in reddit-transductive have been removed). We also investigate 2-layer DP-GCNs in this setting. Once the DP-GNN parameters have been learnt privately over the training graph, we assume that the test graph and test nodes are available non-privately to the inference algorithm.  
\autoref{tab:inductive_results} presents accuracy of our DP-GNN method with 1-layer and 2-layer GCN models on three datasets. We observe that both 1-layer and 2-layer DP-GCNs are significantly more accurate than MLP and DP-MLP models which completely ignore the graph features.

\begin{table}[!htbp]
\caption{\textbf{Test performance of DP-GCN in the inductive setting, with privacy budget $\epsilon \leq 15$.}} \vspace*{-8pt}
\centering
\resizebox{0.8\linewidth}{!}{\begin{tabular}{cccc}
\hline
Model & ogbn-arxiv-disjoint & ogbn-arxiv-clustered & reddit-disjoint \\
\hline \hline
GCN (2-layer)             & 60.641 $\pm$ 0.417 &	56.932 $\pm$ 0.864 &	93.775 $\pm$ 0.116 \\
GCN (1-layer)             & 60.570 $\pm$ 0.438 &	54.188 $\pm$ 0.761 &	92.358 $\pm$ 0.815 \\
\hline
\textbf{DP-GCN (2-layer)} & 55.988 $\pm$ 0.130 &	42.074 $\pm$ 1.054 &	91.234 $\pm$ 0.098 \\
\textbf{DP-GCN (1-layer)} & 56.074 $\pm$ 0.259 &	42.406 $\pm$ 0.902 &	89.551 $\pm$ 0.175 \\
\hline 
MLP                       & 55.460 $\pm$ 0.188 &    37.764 $\pm$ 1.014 &    72.272 $\pm$ 0.132 \\
DP-MLP                    & 52.624 $\pm$ 0.312 &	31.834 $\pm$ 2.801 &	69.998 $\pm$ 0.105 \\
\hline 
\end{tabular}}
\label{tab:inductive_results}
\end{table} 

\subsection{Results with other GNN Architectures}
\label{subsec:other-architectures}

As mentioned in \autoref{sec:main-algorithm}, the DP-GNN training mechanisms can be used with most $r$-layer GNN architectures. We experiment with two more GNN architectures, namely GIN
\citep{xu:gin} and GAT \citep{petar:gat} in both transductive
(\autoref{tab:transductive_results_other_architectures}) and inductive 
(\autoref{tab:inductive_results_other_architectures}) 
settings. 

We observe that DP-GNN performs well across different architectures
in both privacy settings, outperforming MLP and DP-MLP baselines in all cases.
For both the inductive and transductive settings, we observe that GIN performs similarly to GCN, and DP-GIN again has similar performance as DP-GCN. On the ogbn-arxiv-clustered dataset, however, both 1-layer and 2-layer DP-GIN models perform much
better than their DP-GCN counterparts.  

\begin{table}[!htbp]
\caption{\textbf{Test accuracy of DP-GIN and DP-GAT on the transductive ogbn-arxiv dataset with a privacy budget of $\epsilon \leq 12.$}}
\centering
\begin{tabular}{ccc}
\hline
Model & Non-Private GNN & DP-GNN \\
\hline \hline
GCN & 67.758 $\pm$ 0.418 &	61.750 $\pm$ 0.282 \\
GIN & 66.934 $\pm$ 0.529 &	60.859 $\pm$ 0.080 \\
GAT & 65.490 $\pm$ 0.454 &	57.132 $\pm$ 0.116 \\
\hline
MLP & 55.236 $\pm$ 0.317 &	52.178 $\pm$ 0.158 \\
\hline 
\end{tabular}
\label{tab:transductive_results_other_architectures}
\end{table}

\begin{table}[!htbp]
\caption{\textbf{Test performance of DP-GIN in the inductive setting, with privacy budget $\epsilon \leq 15$.}}
\centering
\resizebox{0.8\linewidth}{!}{
\begin{tabular}{cccc}
\hline
Model & ogbn-arxiv-disjoint & ogbn-arxiv-clustered & reddit-disjoint \\
\hline \hline
GIN (2-layer)             & 60.020 $\pm$ 0.697 &	54.239 $\pm$ 1.801 &	93.102 $\pm$ 0.211 \\
GIN (1-layer)             & 59.215 $\pm$ 0.332 &	51.834 $\pm$ 1.362 &	92.682 $\pm$ 0.191 \\
\hline
\textbf{DP-GIN (2-layer)} & 56.396 $\pm$ 0.476 &	45.662 $\pm$ 0.353 &	90.294 $\pm$ 0.484 \\
\textbf{DP-GIN (1-layer)} & 56.914 $\pm$ 0.403 &	43.243 $\pm$ 1.607 &	90.382 $\pm$ 0.065 \\
\hline 
MLP                       & 55.460 $\pm$ 0.188 &	37.764 $\pm$ 1.014 &	72.272 $\pm$ 0.132 \\
DP-MLP                    & 52.624 $\pm$ 0.312 &	31.834 $\pm$ 2.801 &	69.998 $\pm$ 0.105 \\
\hline 
\end{tabular}
}
\label{tab:inductive_results_other_architectures}
\end{table}

\subsection{Ablation Studies}
\label{sec:batch_size_ablation}

{\bf Batch size $m$}: As has been noted in other DP-SGD works \citep{abadi:dp-sgd,bagdasaryan2019differential,mcmahan:tfp},
we observe that increasing the batch size improves the
performance of the learnt DP-GNN model.
As described in \autoref{app:reproducibility}, we find that a batch size of $5000$ to $10000$ to work reasonably well for training DP-GNNs across datasets.

\begin{table}[!t]
\parbox{.45\linewidth}{
\caption{\textbf{GCN and DP-GCN on the ogbn-arxiv dataset with different batch sizes, with DP-GCN privacy budget as $\epsilon \leq 12$.}}
\vspace*{-10pt}
\centering
\resizebox{0.97\linewidth}{!}{\begin{tabular}{cccc}
\hline
Batch Size & GCN $(A_\text{GCN})$ & DP-GCN $(A_\text{DP-GCN})$ & $A_\text{GCN} - A_\text{DP-GCN}$ \\
\hline \hline
2500       &	68.809 &	53.514 &	15.295 \\
5000       &	68.577 &	59.893 &	8.684  \\
10000      &	68.562 &	62.343 &	6.219  \\
20000      &    68.393 &	62.995 &	5.398  \\
40000      &	68.208 &	63.430 &	4.778  \\
Full-Batch &	68.047 &	63.662 &	4.385  \\
\hline
\end{tabular}}
\label{tab:arxiv_batch_size_ablation}
}
\hfill
\parbox{.45\linewidth}{
\caption{\textbf{GCN and DP-GCN on the ogbn-arxiv dataset with different in-degrees, with DP-GCN privacy budget as $\epsilon$ $\leq 12$.}} 
\vspace*{-10pt}
\centering
\resizebox{0.97\linewidth}{!}{\begin{tabular}{cccc}
\hline
Degree & GCN  $(A_\text{GCN})$ & DP-GNN  $(A_\text{DP-GCN})$ & $A_\text{GCN} - A_\text{DP-GCN}$ \\
\hline \hline
3  &	67.931 &	62.413 &	5.518 \\
5  &	68.003 &	62.830 &	5.173 \\
7  &	67.622 &	62.884 &	4.738 \\
10 &	67.615 &	62.534 &	5.081 \\
20 &	68.185 &	61.506 &	6.679 \\
30 &	68.241 &	60.528 &	7.713 \\
\hline 
\end{tabular}}
\label{tab:arxiv_degree_ablation}
}
\end{table}

\begin{figure}[t]
\centering
    \begin{subfigure}[t]{.45\columnwidth}
    	\centering
        \includegraphics[width=\textwidth]{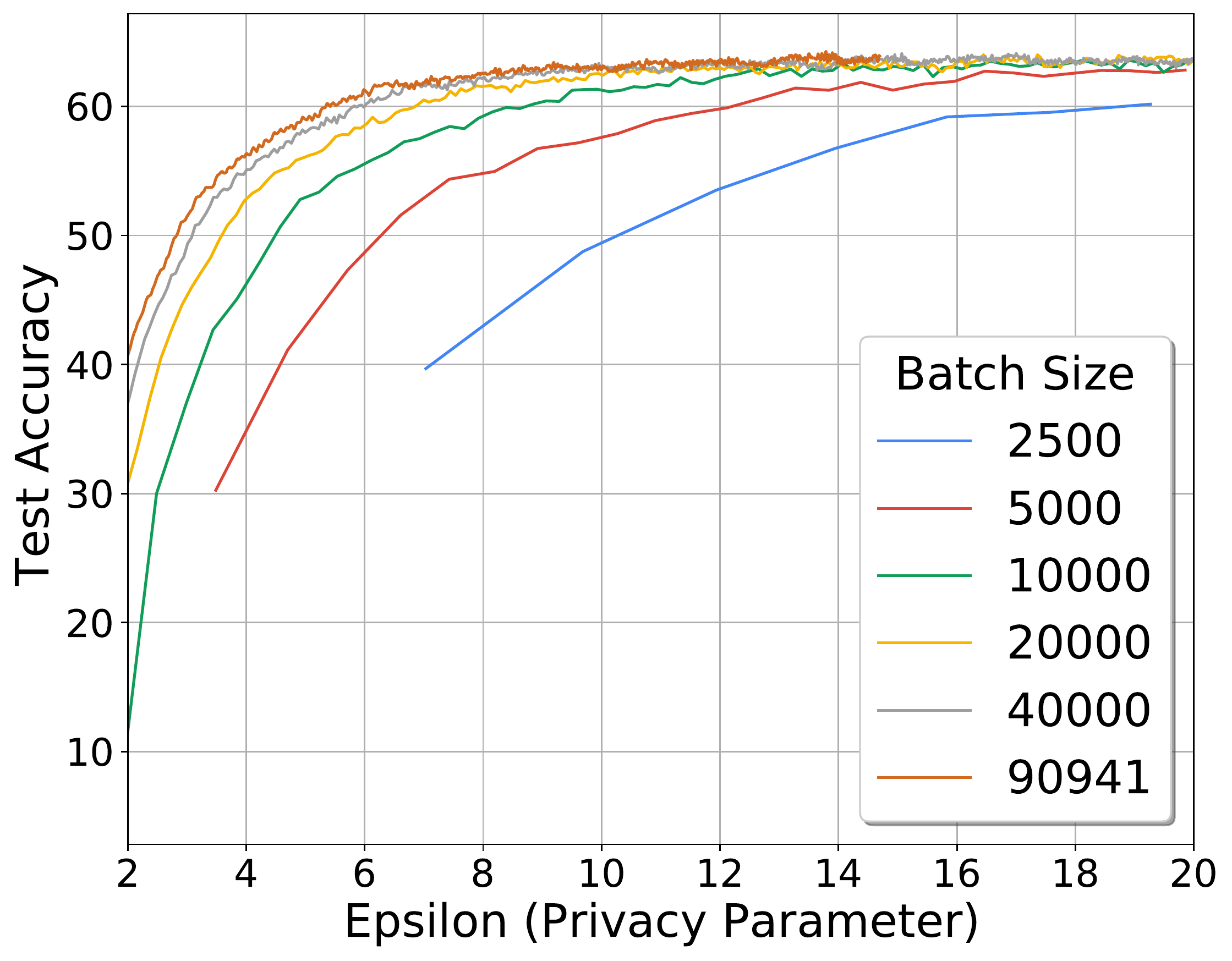}\vspace*{-5pt}
    	\caption{Varying Batch Size $m$}\label{fig:arxiv_batch_size_ablation}		
    \end{subfigure}
    \begin{subfigure}[t]{.46\columnwidth}
    	\centering
        \includegraphics[width=\textwidth]{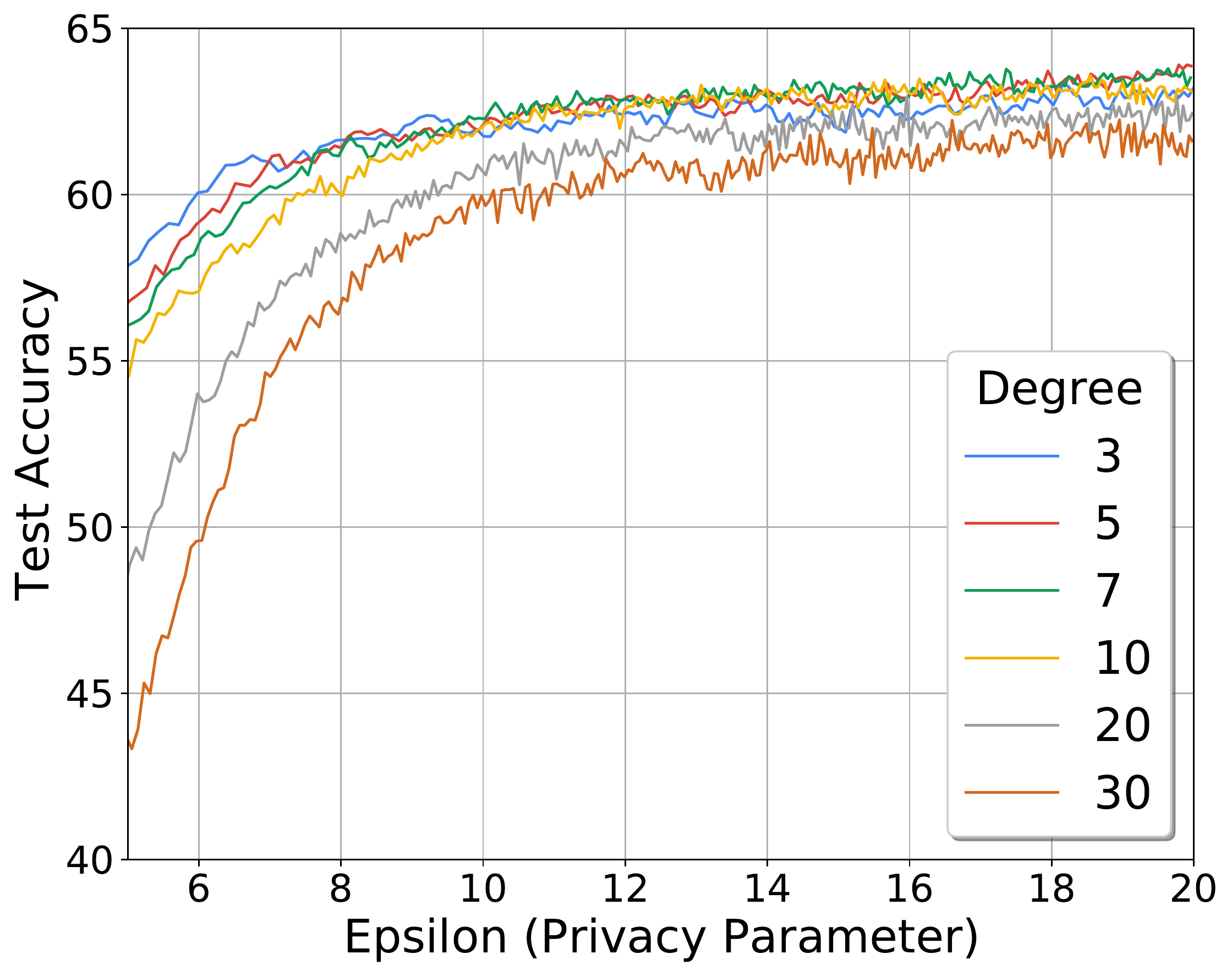}\vspace*{-5pt}
    	\caption{Varying Maximum Degree $K$}\label{fig:arxiv_degree_ablation}
    \end{subfigure}
    \caption{\textbf{Ablation studies on DP-GCN on the ogbn-arxiv dataset.} (a) shows privacy-utility curves for different batch sizes of DP-GCN, such that the scale of the DP noise added per update step is the same. (b) shows privacy-utility curves for varying maximum in-degree $K$ for the {DP-GCN}.
    In both analyses, the other hyperparameters are kept fixed.}\label{fig:arxiv_ablations}
\end{figure}


{\bf Maximum In-Degree $K$}: 
\label{sec:degree_ablation}
Compared to the batch size $m$, the maximum in-degree $K$ has less of an effect on both non-private and private models trained on ogbn-arxiv, as \autoref{tab:arxiv_degree_ablation} shows.
There is still a trade-off: a smaller degree $K$ means lesser differentially private noise added at each update step, but also fewer neighbours for each node to aggregate information from. As described in \autoref{app:reproducibility}, we find that a maximum degree $K$ of around $10$ to work reasonably well across datasets.

%% file: future-work.tex
\section{Conclusions and Future Work}
In this work, we proposed a method to privately learn multi-layer GNN parameters,
that outperforms both private and non-private baselines that do not utilize
graph information. Our method ensures node-level differential privacy, by a careful combination of sensitivity analysis of the gradients and a privacy amplification result extended to the GNN setting. We believe that our work is a first step in the direction of designing powerful GNNs while preserving privacy.
Some promising avenues for future work include extending the DP-GNN method to learn non-local GNNs, addressing fairness issues,
and understanding utility bounds for GNNs with node-level privacy. 
\label{sec:conclusion}

%% file: appendix.tex
\section*{Appendix}

\input{proofs}

\input{classwise}
\input{dp-adam}
\input{reproducibility}

%% file: proofs.tex
\section{Proof that $\mathsf{SAMPLE-SUBGRAPHS}$ Satisfies Node Occurrence Constraints}
\label{app:sampling-proof}

For clarity, we restate \autoref{lm:sampling} below.
\begin{lemma*}
Let $G$ be any graph
with set of training nodes $V_{tr}$.
Then, for any $K, r \geq 0$, the number of occurrences of any node
in the set of training subgraphs
$\mathsf{SAMPLE-SUBGRAPHS}(G, V_{tr}, K, r)$
is bounded above by $N(K, r)$, where:
$$
N(K, r) = \sum_{i = 0}^r K^i = \frac{K^{r + 1} - 1}{K - 1} \in \Theta(K^r)$$
\end{lemma*}

\begin{proof}
Fix any $K \geq 0$. We proceed by induction on $r$.
Note proof for $r = 0$, where $S_v$ is simply $\{v\}$,
is obvious. In this case,
every node $v$ occurs only in its own subgraph $S_v$, and hence, $N(K, r) = \frac{K^{1} - 1}{K - 1} = 1$. For any node $v$ and any $r \geq 0$, let $\calS^{r}(v)$
be the set of all subgraphs in
$\mathsf{SAMPLE-SUBGRAPHS}(G, V_{tr}, K, r)$
in which $v$ occurs.
Then, the hypothesis is that $|\calS^{r}(v)| \leq N(K, r)$
for any $v$.

Assume that the given hypothesis holds for some $r$.
We will now show that our hypothesis holds for $r + 1$ as well,
proving our claim via induction for all $r \geq 0$.

Fix a node $v \in V$.
Then, from the definition of
$\mathsf{DFS-TREE}$ (\autoref{alg:dfs-tree}),
if $S_{u'} \in S^{r + 1}(v)$,
then there must exist $u$ such that
$u \in \bfE_{u'}$ and $S_{u'} \in \calS^{r}(v)$.

By the guarantee of
$\mathsf{SAMPLE-EDGELISTS}$
(\autoref{alg:sample-edgelists}),
the number of nodes such that
$u \in \bfE_{u'}$ is atmost $K$,
any node $u$ is present in atmost $K$ edgelists,
since its (sampled) in-degree is bounded by $K$.
By the inductive hypothesis, there are atmost
$N(K, r) - 1$ such nodes
$u$ such that $S_{u'} \in \calS^{r}(v)$.

Combining the two upper bounds,
and including the subgraph $S_v$ (to which $v$ always belongs),
we can derive the upper bound matching the inductive hypothesis for $r + 1$:
\begin{align*}
    |\calS^{r + 1}(v)| \leq N(K, r + 1) &= K \cdot (N(K, r) - 1) + 1
    = \frac{K^{r + 2} - 1}{K - 1}.
\end{align*}
\end{proof}

\section{Proof of
Privacy Amplification by Subsampling Result for DP-GNN}
\label{app:gnn-amplified-privacy-guarantee-proof}

We provide a detailed proof for \autoref{thm:gnn-amplified-privacy-guarantee} in this section.

\begin{lemma}[Node-Level Sensitivity of any $r$-Layer GNN]
\label{lm:gnn-sensitivity}
Let $G$ be any graph such that 
the maximum in-degree of $G$ is bounded by $K \geq 0$.
Let $V_{tr}$ be the training set of nodes.
Let $\calB_t$ be any choice of $m$ unique subgraphs from
$\calS_{tr} = \mathsf{SAMPLE-SUBGRAPHS}(G, V_{tr}, K, r)$.
For each node $v \in V_{tr}$, let $\hat{y}_v$
be the prediction from an $r$-layer $\mathsf{GNN}$
when run on the subgraph $S_v \in \calS_{tr}$.
Now, $
    \hbfy_v := \mathsf{GNN}(S_v,\bfX,v; \bfTheta)$. 
Consider the loss function $\calL$ of the form: $    \calL(G, {\bf\Theta}) = \sum_{v \in V} \ell\left(\hbfy_v;\bfy_v\right).$
Consider the following quantity $\bfu_t$ from \autoref{alg:dp-sgd}:
\begin{align*}
    \bfu_t(G) &=
    \sum_{v \in \calB_t} \ClipC{\dW \ell\left(\hbfy_v ;\bfy_v\right)}
\end{align*}
Then, the following inequality holds: $$\Delta_K(\bfu_t) < 2C \cdot N(K, r) = 2C \cdot \frac{K^{r + 1} - 1}{K - 1}.$$
\end{lemma}
\begin{proof}
\label{proof:1-layer-sensitivity}
Let $G$ be any graph such that 
the in-degrees of all nodes in $G$ are bounded by $K \geq 0$.
Let $V_{tr}$ be the training set of nodes.
Consider a graph $G'$ formed by removing a
single node $\hbfv$ from $G$,
so that $G$ and $G'$ are node-level adjacent.

Let us define the two sets of subgraphs from $G$ and $G'$ as follows:
\begin{align*}
    \calS_{tr}  &= \mathsf{SAMPLE-SUBGRAPHS}(G, V_{tr}, K, r) \\
    \calS_{tr}' &= \mathsf{SAMPLE-SUBGRAPHS}(G', V_{tr}, K, r)
\end{align*}

Then, the only subgraphs which have changed between
$\calS_{tr}$ and $\calS_{tr}'$ are
only those subgraphs in which $\hbfv$ occurred in,
that is, $\calS^{r}(v)$ from the proof of \autoref{lm:sampling}.
Further, from \autoref{lm:sampling},
there are atmost $N(K, r) = \frac{K^{r + 1} - 1}{K - 1}$ such subgraphs.

We wish to bound the $\ell^2$-norm of the following quantity:
\begin{align*}
    \bfu_t(G)
    -
    \bfu_t(G')
\end{align*}
For convenience, for any node $v$,
we denote the corresponding gradient terms $\dW \ell_v$ and $\dW \ell_v'$ 
as:
\begin{align*}
    \dW \ell_v =
    \dW \ell\left(\mathsf{GNN}(S_v,\bfX,v; \bfTheta); \bfy_v\right)
    &=
    \dW \ell\left(\hbfy_v; \bfy_v\right)
    \\
    \dW \ell_v' =
    \dW \ell\left(\mathsf{GNN}(S_v',\bfX',v; \bfTheta); \bfy_v\right)
    &=
    \dW \ell\left(\hbfy_v'; \bfy_v\right)
\end{align*}

Thus, it is clear that the only gradient
terms $\dW \ell_v$ affected when adding or removing node $\hbfv$,
are those corresponding to the subgraphs in $\calS^{r}(v)$:
\begin{align*}
    \bfu_t(G) - \bfu_t(G')
    = \sum_{\substack{S_v \in (\calB_t \ \cap \ \calS^{r}(v))}}
    \ClipC{\dW \ell_v}  - \ClipC{\dW \ell_v'}
\end{align*}

In the worst case, all of the subgraphs in $\calS^{r}(v)$ 
occur in $\calB_t$.
Each of the gradient terms are clipped to have $\ell^2$-norm $C$.
Hence, the triangle inequality gives us:
\begin{align*}
  \normF{\ClipC{\dW \ell_v}  - \ClipC{\dW \ell_v'}}    \leq \normF{\ClipC{\dW \ell_v}}
  + \normF{\ClipC{\dW \ell_v}}
  = 2C.
\end{align*}
Thus:
\begin{align*}
    \normF{\bfu_t(G) - \bfu_t(G')} \leq 2C \cdot N(K, r) =
    2C \cdot \frac{K^{r + 1} - 1}{K - 1}.
\end{align*}
The same reasoning applies to the case
when $G'$ is formed by a addition of a new node
$\hbfv$ to the graph $G$.
As $G$ and $G'$ were an arbitrary pair of node-level adjacent
graphs, the proof is complete. 
\end{proof}

\begin{lemma}[Un-amplified Privacy Guarantee for Each Iteration of \autoref{alg:dp-sgd}]
\label{thm:dp-sgd-privacy}
Every iteration $t$ of \autoref{alg:dp-sgd}
is $(\alpha, \gamma)$ node-level R\'enyi DP
where $\gamma = \frac{\alpha\cdot(\Delta_K(\bfu_t))^2}{2\sigma^2}$.
\end{lemma}
\begin{proof}
Follows directly from \citet[Corollary 3]{mironov2017renyi}.
\end{proof}

\begin{lemma}[Distribution of Loss Terms Per Minibatch]
\label{lm:distribution-batch}
For any iteration $t$ in
\autoref{alg:dp-sgd}, consider the minibatch $\calB_t$ of
subgraphs. For any subset $\calS$ of $d$ unique subgraphs, define the random variable $\rho$
as $|\calS \cap \calB_t|$.
Then, the distribution of $\rho$ follows the hypergeometric
distribution
$\textup{Hypergeometric}(N, d, m)$:
\begin{align*}
    \rho_i = P[\rho = i] = \frac{\binom{d}{i}\binom{N - d}{m - i}}{\binom{N}{m}}, 
\end{align*}
where $N$ is the number of nodes in the training set
$V_{tr}$ and
$|\calB_t| = m$ is the batch size.
\end{lemma}

\begin{lemma}[Adaptation of Lemma 25 from \citet{FMTT18}]
\label{lm:mixture-rdp}
Let $\mu_0, \ldots, \mu_n$ and $\nu_0, \ldots, \nu_n$ be probability distributions over some domain $Z$ such that: $    \Rd{\alpha}{\mu_0}{\nu_0} \leq \epsilon_0, \ \ldots, \
    \Rd{\alpha}{\mu_n}{\nu_n} \leq \epsilon_n$, 
for some given $\epsilon_0, \ldots, \epsilon_n$.

Let $\rho$ be a probability distribution over $[n] = \{0, \ldots, n\}$.
Denote by $\mu_\rho$ (respectively, $\nu_\rho$)
the probability distribution over $Z$ obtained
by sampling $i$ from $\rho$ and then outputting
a random sample from $\mu_i$ (respectively, $\nu_i$). Then:
\begin{align*}
    \Rd{\alpha}{\mu_\rho}{\nu_\rho} \leq
    \ln \E_{i\sim \rho}\left[e^{\epsilon_i(\alpha - 1)}\right] = 
    \frac{1}{\alpha - 1} \ln
    \sum_{i = 0}^n \rho_i e^{\epsilon_i (\alpha - 1)}.
\end{align*}
\end{lemma}

\begin{lemma}
\label{lm:hypergeom-dominance}
Let $\rho$, $\rho'$ be sampled from the hypergeometric
distribution:
$
    \rho \sim \textup{Hypergeometric}(N, k, m) $, $    \rho' \sim \textup{Hypergeometric}(N, k', m),$ 
such that $k \geq k'$. Then, $\rho$ stochastically dominates $\rho'$:
$
    F_{\rho'}(i) \geq F_{\rho}(i) \textup{ for all } i \in \R$, 
where $F_\rho$ (respectively, $F_{\rho'}$) is the cumulative distribution function (CDF) of $\rho$ (respectively, $\rho'$).
\end{lemma}


For clarity, we restate \autoref{thm:gnn-amplified-privacy-guarantee} below.
\begin{theorem*}[Amplified Privacy Guarantee for any $r$-Layer GNN]
Consider the loss function $\calL$ of the form: 
$$
\calL(G, \bfTheta) = \sum_{v \in V_{tr}} \ell\left(\gnnpt;\bfy_v\right).
$$
Recall, $N$ is the number of training nodes $V_{tr}$, $K$ is the maximum in-degree of the
input graph, $r$ is the number of GNN layers,
and $m$ is the batch size.
For any choice of the noise standard
deviation $\sigma > 0$ and clipping threshold $C$, every iteration $t$ of \autoref{alg:dp-sgd}
is $(\alpha, \gamma)$ node-level
R\'enyi DP, where:
\begin{align*}
\begin{split}
\gamma = \frac{1}{\alpha - 1} \ln \E_{\rho}\left[\exp{\left(\alpha (\alpha - 1) \cdot \frac{2\rho^2C^2}{\sigma^2}\right)}\right], \ \rho \sim \textup{Hypergeometric}\left(N, \frac{K^{r + 1} - 1}{K - 1}, m\right).
\end{split}
\end{align*}
$\textup{Hypergeometric}$ denotes the standard hypergeometric distribution \citep{forbes:statistical}. By the standard composition theorem for R\'enyi Differential Privacy~\citep{mironov2017renyi}, over $T$ iterations, \autoref{alg:dp-sgd} is $(\alpha, \gamma T)$ node-level R\'enyi DP, where $\gamma$ and $\alpha$ are defined above.
\end{theorem*}

\begin{proof}[Proof of \autoref{thm:gnn-amplified-privacy-guarantee}]
\label{proof:gnn-amplified-privacy-guarantee}
At a high-level, \autoref{lm:gnn-sensitivity} tells us that
a node can participate in $N(K, r) = \frac{K^{r + 1} - 1}{K - 1}$
training subgraphs from $\calS_{tr}$, in the worst case.
However, on average, only a fraction of these subgraphs
will be sampled in the mini-batch $\calB_{t}$,
as \autoref{lm:distribution-batch} indicates.
We use the knowledge of the exact distribution
of the number of subgraphs sampled in $\calB_{t}$ provided by \autoref{lm:distribution-batch}
with \autoref{lm:mixture-rdp}
to get a tighter bound on the R\'enyi divergence
between the distributions of $\Tilde{\bfu}_t$
over node-level adjacent graphs.
Finally, \autoref{lm:hypergeom-dominance} allows us to make the above bound
independent of the actual node being removed, giving our final result.


Let $G$ be any graph with training set $V_{tr}$. Let $G'$ be formed by removing a
single node $\hbfv$ from $G$,
so  $G$ and $G'$ are node-level adjacent.

For convenience, for any node $v$,
we denote the corresponding gradient terms $\dW \ell_v$ and $\dW \ell_v'$ 
as:
\begin{align*}
    \dW \ell_v =
    \dW \ell\left(\mathsf{GNN}(S_v,\bfX,v; \bfTheta); \bfy_v\right)
    &=
    \dW \ell\left(\hbfy_v; \bfy_v\right)
    \\
    \dW \ell_v' =
    \dW \ell\left(\mathsf{GNN}(S_v',\bfX',v; \bfTheta); \bfy_v\right)
    &=
    \dW \ell\left(\hbfy_v'; \bfy_v\right)
\end{align*}

Let $\calS^{r}(v)$
be the set of all subgraphs in
$\mathsf{SAMPLE-SUBGRAPHS}(G, V_{tr}, K, r)$
in which $v$ occurs.
\begin{align*}
    \bfu_t(G) - \bfu_t(G')
    = \sum_{\substack{S_v \in (\calB_t \ \cap \ \calS^{r}(\hbfv))}}
    \ClipC{\dW \ell_v}  - \ClipC{\dW \ell_v'}.
\end{align*}
Using the notation from \autoref{alg:dp-sgd}, we have:
\begin{align*}
    \Tilde{\bfu}_t(G) &= \bfu_t(G) + \calN(0, \sigma^2 \mathbb{I})
    \\
    \Tilde{\bfu}_t(G') &= \bfu_t(G') + \calN(0, \sigma^2 \mathbb{I})
\end{align*}
We need to show that $
    \Rd{\alpha}{\Tilde{\bfu}_t(G)}{\Tilde{\bfu}_t(G')} \leq \gamma.
$

From the above equation, 
we see that the sensitivity of $\bfu_t$ depends on the number
of subgraphs
in $\calS^{r}(v)$ that are present in $\gB_t$.
Let $\rho'$ be the distribution over $\{0, 1, \ldots |\calS^{r}(v)| \}$ of the number of subgraphs in $\calS^{r}(v)$
present in $\calB_t$, that is, $\rho' = | \calS^{r}(v)\cap \calB_t |$.
\autoref{lm:distribution-batch} then gives us that the distribution of $\rho'$ is:
$
    \rho' \sim \textup{Hypergeometric}(N, \calS^{r}(v), m).
$
In particular, when $\rho' = i$, exactly $i$
subgraphs are
sampled in $\gB_t$.
Then, following
\autoref{lm:gnn-sensitivity}, 
$
    \Delta_K(\bfu_t \ | \ \rho' = i) < 2iC.
$
Thus, conditioning on $\rho' = i$,
we see that every iteration is
$(\alpha, \gamma_i)$
node-level R\'enyi DP, by \autoref{thm:dp-sgd-privacy}
where $
    \gamma_i = \alpha\cdot 2i^2C^2/\sigma^2.
    \label{eqn:gamma-i}
$

Define the distributions $\mu_i$ and $\nu_i$
for each $i \in \{0, \ldots, |\calS^{r}(v)|\}$, as follows:
$
    \mu_i = \left[\Tilde{\bfu}_t(G) \ | \ \rho' = i\right],  
    \nu_i = \left[\Tilde{\bfu}_t(G') \ | \ \rho' = i\right]$.
Then $
    \Rd{\alpha}{\mu_i}{\nu_i} \leq \gamma_i$. For the mixture distributions $\mu_{\rho'} = \Tilde{\bfu}_t(G)$ and
$\nu_{\rho'} = \Tilde{\bfu}_t(G')$,
\autoref{lm:mixture-rdp} now tells us that:
\begin{align*}
    \Rd{\alpha}{\Tilde{\bfu}_t(G)}{\Tilde{\bfu}_t(G')}
    &= \Rd{\alpha}{\mu_{\rho'}}{\nu_{\rho'}}
    \\
    &\leq
    \frac{1}{\alpha - 1} \ln \E_{i\sim \rho'}\left[\exp{\left(\gamma_i(\alpha - 1)\right)}\right]
    \\
    &=
    \frac{1}{\alpha - 1} \ln \E\left[f(\rho')\right].
    \label{eqn:f-rho'-usage}
\end{align*}
where:
\begin{align*}
    f(\rho') = \exp{\left(\alpha(\alpha - 1)\cdot\frac{ 2\rho'^2C^2}{\sigma^2}\right)}.
\end{align*} 
Define another distribution $\rho$ as: $$
    \rho \sim \textup{Hypergeometric}(N, N(K, r), m) $$
where $N(K, r) = \frac{K^{r + 1} - 1}{K - 1}$ is the upper bound on $|\calS^{r}(v)|$
from \autoref{lm:sampling}.
By \autoref{lm:hypergeom-dominance}, $\rho$ stochastically
dominates $\rho'$. As non-decreasing functions preserve
stochastic dominance \citep{hadar:stochastic-dominance}, 
$f(\rho)$ stochastically
dominates $f(\rho')$, and hence $
    \E\left[f(\rho')\right] \leq \E\left[f(\rho)\right].
    \label{eqn:f-rho-usage}
$
It follows that:
\begin{align*}
    \Rd{\alpha}{\Tilde{\bfu}_t(G)}{\Tilde{\bfu}_t(G')}
    &\leq \frac{1}{\alpha - 1} \ln \E\left[f(\rho)\right]
    \\
    &= \frac{1}{\alpha - 1} \ln \E_{\rho}\left[\exp{\left(\alpha (\alpha - 1) \cdot \frac{2\rho^2C^2}{\sigma^2}\right)}\right] = \gamma.
\end{align*}
The theorem now follows from the fact that the above holds for an arbitrary pair of node-level adjacent graphs $G$ and $G'$. 
\end{proof}

%% file: classwise.tex
\section{Class-wise Analysis of Learnt Models}
\label{app:class-wise-analysis}

\begin{figure}[htbp]
\centering
    \begin{subfigure}[t]{.48\textwidth}
    	\centering
        \includegraphics[width=\textwidth]{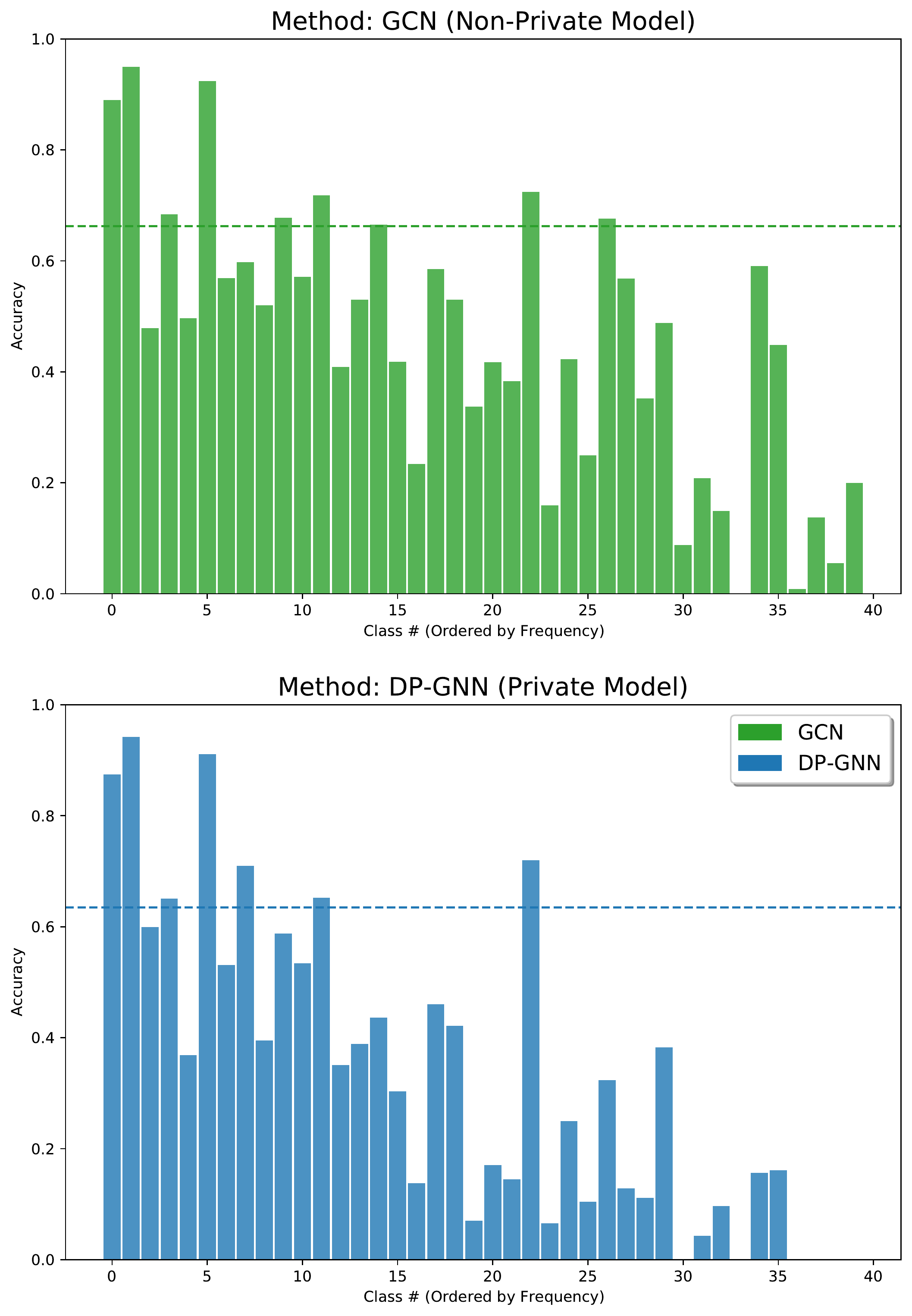}
    	\caption{ogbn-arxiv}\label{fig:arxiv_class_wise}	
    \end{subfigure}
    \quad
    \begin{subfigure}[t]{.48\textwidth}
    	\centering
        \includegraphics[width=\textwidth]{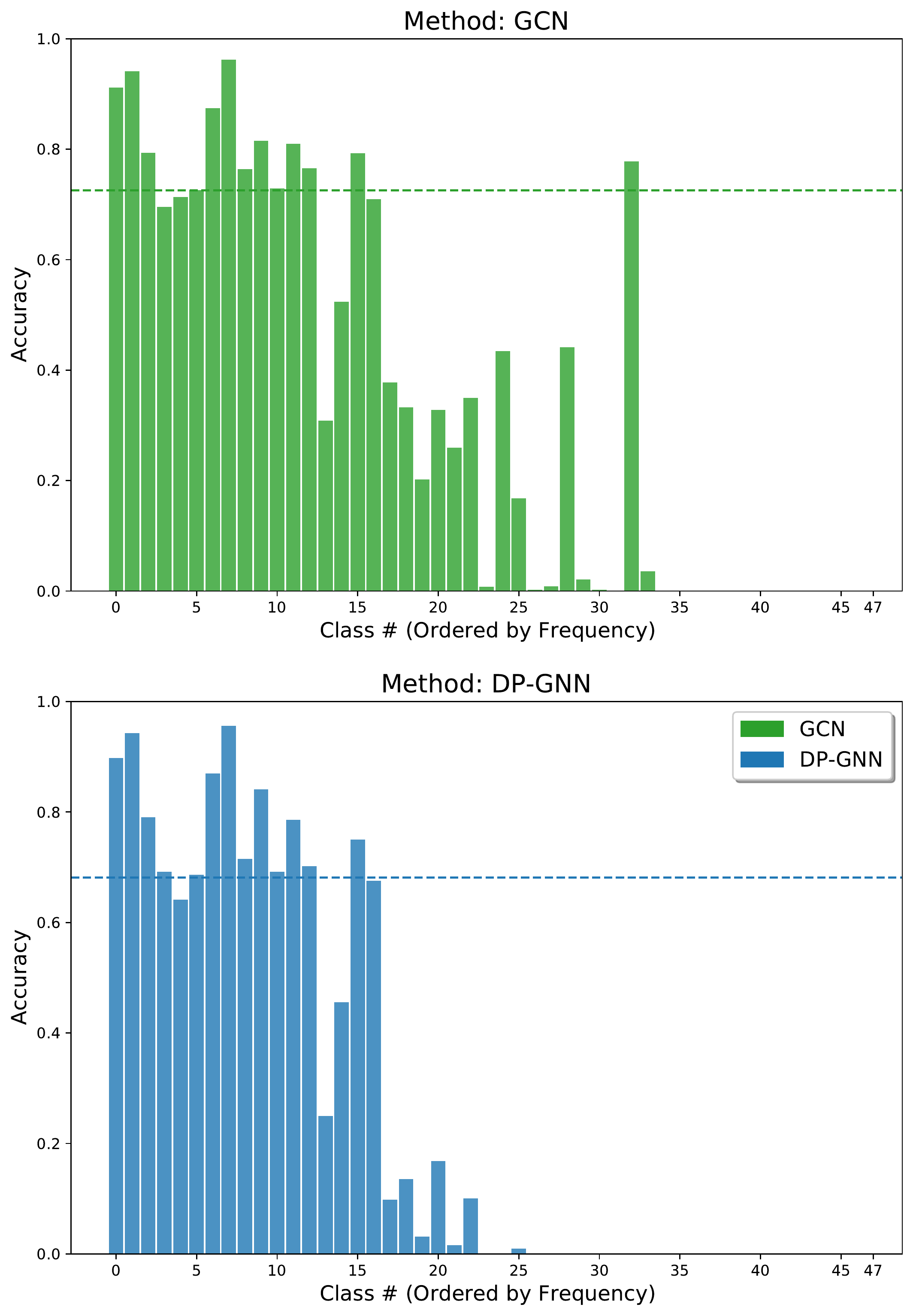}
    	\caption{ogbn-products}\label{fig:products_class_wise}
    \end{subfigure}
    \quad
    \begin{subfigure}[t]{\textwidth}
    	\centering
        \includegraphics[width=.8\textwidth]{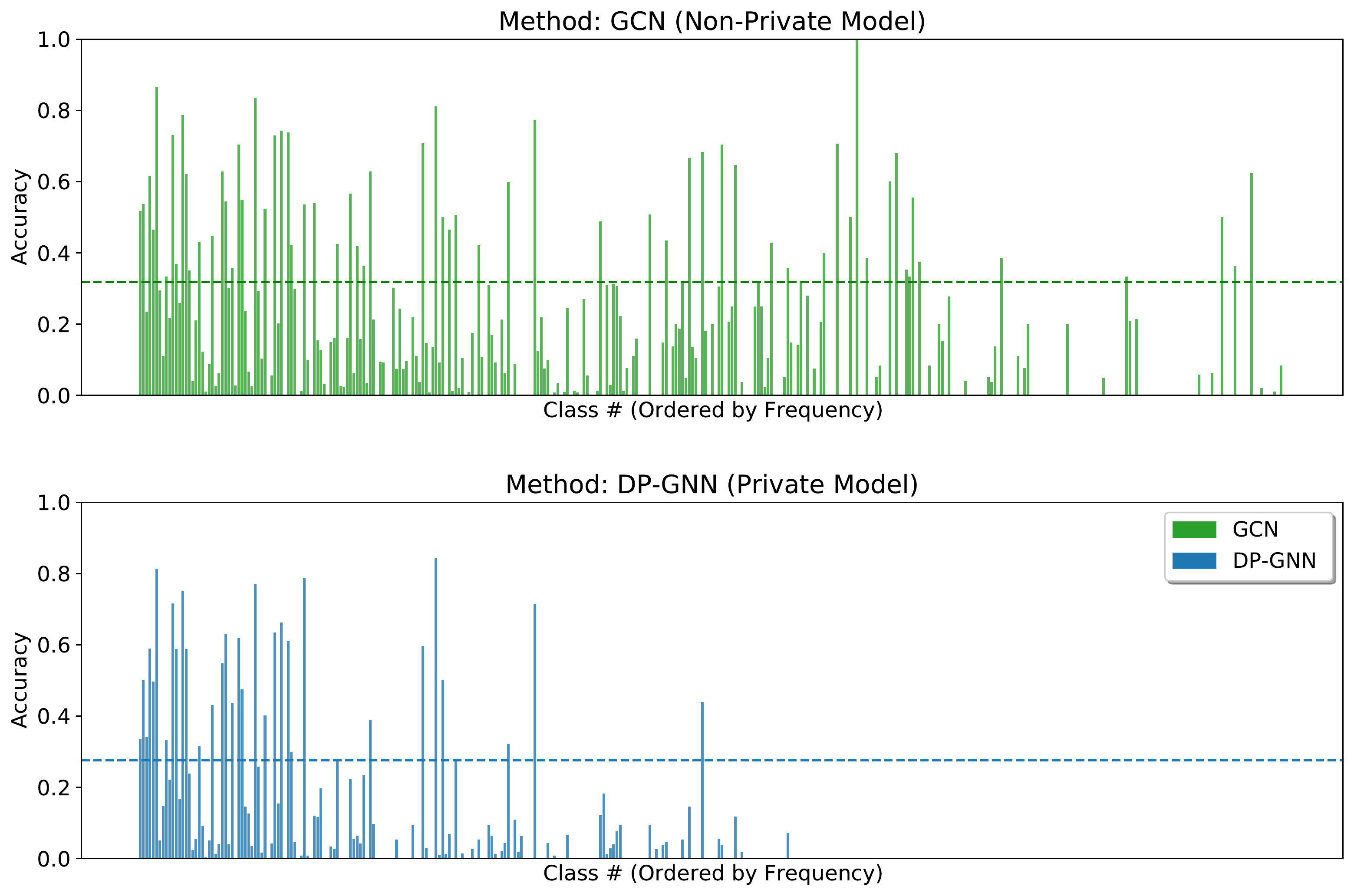}
    	\caption{ogbn-mag}\label{fig:mag_class_wise}
    \end{subfigure}
    \caption{\textbf{Comparison of class-wise test accuracies of the non-private GCN model and private DP-GCN model on all datasets, ordered by the decreasing frequency of occurrence of classes in the training data from left to right.} The dotted lines indicate the overall (`micro') accuracy for each model.}
    \label{fig:supplementary_class_wise_ablations}
\end{figure}

To better understand the performance of the private model as compared to the non-private baseline for our considering setting of multi-class classification at a node-level, we compare the accuracy of these two models for each dataset at a class-wise granularity. These results are illustrated in \autoref{fig:supplementary_class_wise_ablations}. We empirically observe that the performance of the private model degrades as the frequency of training data points for a particular class decreases. This indicates that the model is able to classify data points of `frequent' classes with reasonable accuracy, but struggles with classification accuracy on the data points of `rarer' classes.
This observation is in line with previous
claims from
\citep{bagdasaryan2019differential,fioretto:dp-fairness}
that differentially-private models generally perform
disparately worse on under-represented classes.
While methods \citep{jagielski:dp-fair-learning,xu:dpsgd-f} have been developed
for improving the fairness of DP-SGD,
their extension to the GNN setting represents an important
direction for future work.

\clearpage

%% file: dp-adam.tex
\section{Learning Graph Convolutional Networks (GCN) via DP-Adam}
\label{app:dp-adam}
In \autoref{alg:dp-adam}, we provide the description of DP-Adam,
which adapts \autoref{alg:dp-sgd} to use the popular Adam \citep{kingma2014adam} optimizer, instead of SGD.
The privacy guarantee and accounting for
\autoref{alg:dp-adam} is identical to that of
\autoref{alg:dp-sgd}, since the DP clipping and noise addition
steps are identical.

\begin{algorithm}[htbp]
\caption{DP-GNN (Adam): Differentially Private Graph Neural Network with Adam}
\KwData{Graph $G = (V, E, \bfX, \bfY)$, GNN definition $\mathsf{GNN}$, Training set $V_{tr}$, Loss function $\calL$, Batch size $m$, Maximum degree $K$, Learning rate $\eta$, Clipping threshold $C$, Noise standard deviation $\sigma$, Maximum training iterations $T$, Adam hyperparameters $(\beta_1, \beta_2)$.}
\KwResult{GNN parameters $\bfTheta_{T}$.}
 Note that $V_{tr}$ is the subset of nodes for which labels are
 available (see Paragraph 1 of \autoref{sec:preliminaries}). \\
 Construct the set of training subgraphs with \autoref{alg:sample-subgraphs}: $\calS_{tr} \gets \mathsf{SAMPLE-SUBGRAPHS}(G, V_{tr}, K, r)$. \\ 
 Initialize $\bfTheta_0$ randomly. \\
 \For{$t = 0$ \KwTo $T$}{
  Sample set $\calB_t \subseteq \calS_{tr}$ of size $m$ uniformly at random from all subsets of $\calS_{tr}$.\\
  Compute the update term $\bfu_t$ as the sum of the clipped gradient terms in the mini-batch $\gB_t$:
  \begin{align*}
    \bfu_t \gets \sum_{S_v \in \gB_t}
    \ClipC{\dT\ell\left(\gnnpt;\bfy_v\right)}
  \end{align*}
  Add independent Gaussian noise to the gradient term:
  $\Tilde{\bfu_t} \gets \bfu_t + \calN(0, \sigma^2 \mathbb{I}) $ \\
  Update first and second moment estimators with the noisy gradient, correcting for bias:
  \begin{align*}
      f_t &\gets \beta_1 \cdot f_{t-1} + (1 - \beta_1) \cdot \Tilde{\bfu_t} \\
      s_t &\gets \beta_2 \cdot s_{t-1} + (1 - \beta_2) \cdot (\Tilde{\bfu_t}\odot\Tilde{\bfu_t})  \\
      \widehat{f}_t &\gets \frac{f_t}{1 - \beta_1^t} \\
      \widehat{s}_t &\gets \frac{s_t}{1 - \beta_2^t}
  \end{align*}
  Update the current estimate of the parameters with the noisy estimators:
  \begin{align*}
      \bfTheta_{t + 1} \gets \bfTheta_{t} - \frac{\eta}{m} \frac{\widehat{f}_t}{\sqrt{\widehat{s}_t^2} + \epsilon}
  \end{align*}
 }
 \label{alg:dp-adam}
\end{algorithm}

%% file: reproducibility.tex
\section{Reproducibility}
\label{app:reproducibility}

Our open-sourced pipeline for sampling graph datasets and training DP-GNN and the other baseline models is available at
\url{https://github.com/google-research/google-research/tree/master/differentially_private_gnns}.


\textbf{Data:} \autoref{tab:dataset_statistics} provides details about the node classification benchmark datasets used in this work: ogbn-arxiv, ogbn-products, ogbn-mag\footnote{Obtained from \url{https://ogb.stanford.edu/docs/nodeprop/}.}, and reddit\footnote{Obtained from \url{http://snap.stanford.edu/graphsage/}. Available preprocessed at \href{https://drive.google.com/corp/drive/folders/1rq-H0XUM0BIRW9Pq5P4FMC9Xirpdx6zs}{this Google Drive link}.}. We use
the following `inverse-degree' normalization of the adjacency matrix for all GCN
models:
$\widehat{\bfA} = (d + \mathbb{I})^{-1}(\bfA + \mathbb{I}).$
We use a variant of the original
GAT architecture, utilizing dot-product attention instead of additive
attention, with $10$ attention heads. 
Adam \citep{kingma2014adam} with $\beta_1=0.9$ and $\beta_2=0.999$, and SGD optimizers were used for training all methods for each of the datasets. A latent size of $256$ was used for the encoder,
GNN and decoder layers.
Additionally, the best hyperparameters corresponding to each experiment to reproduce the results in the main paper are reported in the tables below.

$\lr$ refers to the learning rate, $n_\text{enc}$ refers to the number of layers in the encoder MLP, $n_\text{dec}$ refers to the number of layers in the decoder MLP, $\nm$ refers to the noise multiplier, and $K$ refers to the maximum degree.

\begin{table}[!t]
\caption{\textbf{Statistics of datasets used in our experiments.}} 
\centering
\begin{longtable}{cccccc}
\hline 
Dataset & Nodes &  Avg. Degree & Features & Classes & Train/Val/Test Split \\
\hline
ogbn-arxiv & 169,343 & 13.7 & 128 & 40 & 0.54/0.18/0.28\\
ogbn-arxiv-disjoint & 169,343 & 5.5 & 128 & 40 & 0.54/0.18/0.28\\
ogbn-arxiv-clustered\footnotemark & 169,343 & 13.0 & 128 & 40 & 0.67/0.18/0.15 \\
ogbn-products  & 2,449,029 & 50.5 & 100 & 47 & 0.08/0.02/0.90\\
ogbn-mag & 736,389 & 21.7 & 128 & 349 & 0.85/0.09/0.05\\
reddit & 232,965 & 99.6 & 602 & 41 & 0.66/0.10/0.24\\
reddit-disjoint & 232,965 & 54.4 & 602 & 41 & 0.66/0.10/0.24\\
\hline
\end{longtable}
\label{tab:dataset_statistics}
\end{table} 
\footnotetext[6]{Indices indicating the split for each node available at
\href{https://drive.google.com/file/d/1rIJHZEwEWtG3GFq8KmrHmTWgjYFfOiBd/view?usp=sharing}{this Google Drive link}. `0' indicates the train split, `1' indicates the validation split, and `2' indicates the test split.}

\begin{table}[htbp]
\caption{\textbf{Hyperparameters for models in \autoref{tab:transductive_results}.}}
\centering
\resizebox{0.9\columnwidth}{!}{
        \begin{tabular}{cccccccccc}
            \hline
            Model & Dataset & $\lr$ & Batch Size & Activation & $n_\text{enc}$ & $n_\text{dec}$ & $\nm$ & $K$ \\ \hline
            
            \multirow{4}{*}{GCN}
            & ogbn-arxiv & \hfil 0.002 & \hfil 1,000 & \hfil ReLU & \hfil 2 & \hfil 2 & \hfil - & \hfil 30\\
            & ogbn-products & \hfil 0.001 & \hfil 1,000 & \hfil ReLU & \hfil 2 & \hfil 1 & \hfil - & \hfil 30\\
            & ogbn-mag & \hfil 0.001 & \hfil 1,000 & \hfil ReLU & \hfil 2 & \hfil 2 & \hfil - & \hfil 10\\
            & reddit & \hfil 0.001 & \hfil 1,000 & \hfil ReLU & \hfil 2 & \hfil 2 & \hfil - & \hfil 10\\

            \hline
            
            \multirow{4}{*}{DP-GCN (Adam)}
            & ogbn-arxiv & \hfil 0.003 & \hfil 10,000 & \hfil Tanh & \hfil 1 & \hfil 2 & \hfil 2 & \hfil 7\\
            & ogbn-products & \hfil 0.005 & \hfil 10,000 & \hfil Tanh & \hfil 1 & \hfil 2 & \hfil 1 & \hfil 10\\
            & ogbn-mag & \hfil 0.001 & \hfil 10,000 & \hfil Tanh & \hfil 2 & \hfil 2 & \hfil 2 & \hfil 10\\
            & reddit & \hfil 0.002 & \hfil 10,000 & \hfil Tanh & \hfil 2 & \hfil 2 & \hfil 1 & \hfil 10\\
            \hline
            
            \multirow{4}{*}{DP-GCN (SGD)}
            & ogbn-arxiv & \hfil 1.0 & \hfil 10,000 & \hfil Tanh & \hfil 2 & \hfil 1 & \hfil 2 & \hfil 7\\
            & ogbn-products & \hfil 2.0 & \hfil 40,000 & \hfil Tanh & \hfil 1 & \hfil 1 & \hfil 2 & \hfil 5\\
            & ogbn-mag & \hfil 1.0 & \hfil 10,000 & \hfil ReLU & \hfil 1 & \hfil 2 & \hfil 1 & \hfil 10\\
            & reddit & \hfil 0.1 & \hfil 10,000 & \hfil Tanh & \hfil 1 & \hfil 2 & \hfil 1 & \hfil 10\\
            \hline
            
            \multirow{4}{*}{MLP}
            & ogbn-arxiv & \hfil 0.001 & \hfil 1,000 & \hfil ReLU & \hfil 2 & \hfil 1 & \hfil - & \hfil -\\
            & ogbn-products & \hfil 0.001 & \hfil 1,000 & \hfil ReLU & \hfil 1 & \hfil 2 & \hfil - & \hfil -\\
            & ogbn-mag & \hfil 0.01 & \hfil 1,024 & \hfil ReLU & \hfil 2 & \hfil 1 & \hfil - & \hfil -\\
            & reddit & \hfil 0.001 & \hfil 1,000 & \hfil ReLU & \hfil 1 & \hfil 1 & \hfil - & \hfil -\\
            \hline
            
            \multirow{3}{*}{DP-MLP}
            & ogbn-arxiv & \hfil 0.003 & \hfil 10,000 & \hfil Tanh & \hfil 1 & \hfil 2 & \hfil 1 & \hfil 0\\
            & ogbn-products & \hfil 0.002 & \hfil 10,000 & \hfil ReLU & \hfil 1 & \hfil 2 & \hfil 1 & \hfil 10\\
            & ogbn-mag & \hfil 0.001 & \hfil 10,000 & \hfil ReLU & \hfil 2 & \hfil 2 & \hfil 1 & \hfil 10\\
            & reddit & \hfil 0.001 & \hfil 10,000 & \hfil Tanh & \hfil 1 & \hfil 2 & \hfil 1 & \hfil 10\\
            \hline
        \end{tabular}
        }\vspace*{-5pt}
    \label{tab:transductive_results_best_hyperparams}
\end{table}

\begin{table}[!htbp]
\caption{\textbf{Hyperparameters for models in \autoref{tab:inductive_results}.}}
\centering
\resizebox{\columnwidth}{!}{
        \begin{tabular}{cccccccccc}
            \hline
            Model & Dataset & $\lr$ & Batch Size & Activation & $n_\text{enc}$ & $n_\text{dec}$ & $\nm$ & $K$ \\ \hline
            
            \multirow{3}{*}{GCN (2-layer)}
            & ogbn-arxiv-disjoint & \hfil 0.001 & \hfil 1,000 & \hfil ReLU & \hfil 1 & \hfil 1 & \hfil - & \hfil 30\\
            & ogbn-arxiv-clustered & \hfil 0.001 & \hfil 1,000 & \hfil ReLU & \hfil 1 & \hfil 2 & \hfil - & \hfil 10\\
            & reddit-disjoint & \hfil 0.001 & \hfil 1,000 & \hfil ReLU & \hfil 2 & \hfil 2 & \hfil - & \hfil 10\\
             \hline
            
            \multirow{3}{*}{GCN (1-layer)}
            & ogbn-arxiv-disjoint & \hfil 0.001 & \hfil 1,000 & \hfil ReLU & \hfil 1 & \hfil 2 & \hfil - & \hfil 7\\
            & ogbn-arxiv-clustered & \hfil 0.001 & \hfil 1,000 & \hfil ReLU & \hfil 1 & \hfil 1 & \hfil - & \hfil 30\\
            & reddit-disjoint & \hfil 0.001 & \hfil 1,000 & \hfil ReLU & \hfil 2 & \hfil 2 & \hfil - & \hfil 10\\
            \hline
            
            \multirow{3}{*}{DP-GCN (2-layer)} 
            & ogbn-arxiv-disjoint & \hfil 0.003 & \hfil 20,000 & \hfil Tanh & \hfil 1 & \hfil 1 & \hfil 2 & \hfil 3\\
            & ogbn-arxiv-clustered & \hfil 0.002 & \hfil 20,000 & \hfil Tanh & \hfil 2 & \hfil 1 & \hfil 4 & \hfil 3\\
            & reddit-disjoint & \hfil 0.002 & \hfil 20,000 & \hfil Tanh & \hfil 1 & \hfil 1 & \hfil 2 & \hfil 3\\
            \hline

            \multirow{3}{*}{DP-GCN (1-layer)} 
            & ogbn-arxiv-disjoint & \hfil 0.002 & \hfil 20,000 & \hfil Tanh & \hfil 1 & \hfil 2 & \hfil 4 & \hfil 7\\
            & ogbn-arxiv-clustered & \hfil 0.002 & \hfil 20,000 & \hfil Tanh & \hfil 2 & \hfil 2 & \hfil 4 & \hfil 10\\
            & reddit-disjoint & \hfil 0.001 & \hfil 10,000 & \hfil Tanh & \hfil 2 & \hfil 2 & \hfil 1 & \hfil 10\\
            \hline

            \multirow{3}{*}{MLP} 
            & ogbn-arxiv-disjoint & \hfil 0.001 & \hfil 1,000 & \hfil ReLU & \hfil 2 & \hfil 1 & \hfil - & \hfil -\\
            & ogbn-arxiv-clustered & \hfil 0.002 & \hfil 1,000 & \hfil ReLU & \hfil 1 & \hfil 2 & \hfil - & \hfil -\\
            & reddit-disjoint & \hfil 0.001 & \hfil 1,000 & \hfil ReLU & \hfil 1 & \hfil 1 & \hfil - & \hfil -\\
            \hline
            
            \multirow{3}{*}{DP-MLP}
            & ogbn-arxiv-disjoint & \hfil 0.003 & \hfil 10,000 & \hfil Tanh & \hfil 1 & \hfil 2 & \hfil 1 & \hfil 10\\
            & ogbn-arxiv-clustered & \hfil 0.003 & \hfil 10,000 & \hfil Tanh & \hfil 2 & \hfil 2 & \hfil 1 & \hfil 10\\
            & reddit-disjoint & \hfil 0.001 & \hfil 10,000 & \hfil Tanh & \hfil 1 & \hfil 2 & \hfil 1 & \hfil 10\\
            \hline
        \end{tabular}
        }\vspace*{-5pt}
    \label{tab:inductive_results_best_hyperparams}
\end{table}

\begin{table}[!t]
\caption{\textbf{Hyperparameters for models in \autoref{tab:transductive_results_other_architectures}.}}
\centering
\resizebox{0.9\columnwidth}{!}{
        \begin{tabular}{cccccccccc}
            \hline
            Architecture & Method & $\lr$ & Batch Size & Activation & $n_\text{enc}$ & $n_\text{dec}$ & $\nm$ & $K$ \\ \hline

            \multirow{2}{*}{GCN} & Non-Private GNN & \hfil 0.002 & \hfil 1,000 & \hfil ReLU & \hfil 2 & \hfil 2 & \hfil - & \hfil 30\\
            & DP-GNN & \hfil 0.003 & \hfil 10,000 & \hfil Tanh & \hfil 1 & \hfil 2 & \hfil 2 & \hfil 7\\ \hline
            
            \multirow{2}{*}{GIN} & Non-Private GNN & \hfil 0.003 & \hfil 1,000 & \hfil ReLU & \hfil 2 & \hfil 1 & \hfil - & \hfil 30\\
            & DP-GNN & \hfil 0.003 & \hfil 10,000 & \hfil Tanh & \hfil 1 & \hfil 1 & \hfil 1 & \hfil 7\\ \hline
            
            \multirow{2}{*}{GAT} & Non-Private GNN & \hfil 0.001 & \hfil 1,000 & \hfil Tanh & \hfil 2 & \hfil 2 & \hfil - & \hfil 30\\
            & DP-GNN & \hfil 0.004 & \hfil 20,000 & \hfil Tanh & \hfil 1 & \hfil 2 & \hfil 2 & \hfil 7\\\hline
            
            \multirow{2}{*}{MLP} & Non-Private MLP & \hfil 0.001 & \hfil 1,000 & \hfil ReLU & \hfil 2 & \hfil 1 & \hfil - & \hfil -\\
            & DP-MLP  & \hfil 0.003 & \hfil 10,000 & \hfil Tanh & \hfil 1 & \hfil 2 & \hfil 1 & \hfil -\\\hline
        \end{tabular}
        }\vspace*{-5pt}
    \label{tab:transductive_results_other_architectures_best_hyperparams}
\end{table}

\begin{table}[!t]
\caption{\textbf{Hyperparameters for models in \autoref{tab:inductive_results_other_architectures}.}}
\centering
\resizebox{0.9\columnwidth}{!}{
        \begin{tabular}{cccccccccc}
            \hline
            Model & Dataset & $\lr$ & Batch Size & Activation & $n_\text{enc}$ & $n_\text{dec}$ & $\nm$ & $K$ \\ \hline
            
            \multirow{3}{*}{GIN (2-layer)}
            & ogbn-arxiv-disjoint & \hfil 0.001 & \hfil 1,000 & \hfil ReLU & \hfil 1 & \hfil 2 & \hfil - & \hfil 3\\
            & ogbn-arxiv-clustered & \hfil 0.001 & \hfil 1,000 & \hfil ReLU & \hfil 2 & \hfil 2 & \hfil - & \hfil 30\\
            & reddit-disjoint & \hfil 0.001 & \hfil 1,000 & \hfil ReLU & \hfil 2 & \hfil 2 & \hfil - & \hfil 10\\ \hline
                        
            \multirow{3}{*}{GIN (1-layer)}
            & ogbn-arxiv-disjoint & \hfil 0.002 & \hfil 1,000 & \hfil ReLU & \hfil 2 & \hfil 1 & \hfil - & \hfil 30\\
            & ogbn-arxiv-clustered & \hfil 0.001 & \hfil 1,000 & \hfil ReLU & \hfil 2 & \hfil 1 & \hfil - & \hfil 30\\
            & reddit-disjoint & \hfil 0.001 & \hfil 1,000 & \hfil ReLU & \hfil 2 & \hfil 1 & \hfil - & \hfil 10\\
            \hline
            
            \multirow{3}{*}{DP-GIN (2-layer)} 
            & ogbn-arxiv-disjoint & \hfil 0.002 & \hfil 20,000 & \hfil Tanh & \hfil 2 & \hfil 2 & \hfil 2 & \hfil 3\\
            & ogbn-arxiv-clustered & \hfil 0.002 & \hfil 10,000 & \hfil Tanh & \hfil 1 & \hfil 2 & \hfil 1 & \hfil 3\\
            & reddit-disjoint & \hfil 0.001 & \hfil 10,000 & \hfil Tanh & \hfil 1 & \hfil 1 & \hfil 1 & \hfil 3\\\hline

            \multirow{3}{*}{DP-GIN (1-layer)} 
            & ogbn-arxiv-disjoint & \hfil 0.004 & \hfil 20,000 & \hfil Tanh & \hfil 1 & \hfil 2 & \hfil 2 & \hfil 7\\
            & ogbn-arxiv-clustered & \hfil 0.004 & \hfil 10,000 & \hfil Tanh & \hfil 1 & \hfil 1 & \hfil 1 & \hfil 10\\
            & reddit-disjoint & \hfil 0.001 & \hfil 10,000 & \hfil Tanh & \hfil 2 & \hfil 2 & \hfil 1 & \hfil 10\\
             \hline

            \multirow{3}{*}{MLP} 
            & ogbn-arxiv-disjoint & \hfil 0.001 & \hfil 1,000 & \hfil ReLU & \hfil 2 & \hfil 1 & \hfil - & \hfil 10\\
            & ogbn-arxiv-clustered & \hfil 0.002 & \hfil 1,000 & \hfil ReLU & \hfil 1 & \hfil 2 & \hfil - & \hfil 10\\
            & reddit-disjoint & \hfil 0.001 & \hfil 1,000 & \hfil ReLU & \hfil 1 & \hfil 1 & \hfil - & \hfil 10\\
            \hline
            
            \multirow{3}{*}{DP-MLP}
            & ogbn-arxiv-disjoint & \hfil 0.003 & \hfil 10,000 & \hfil Tanh & \hfil 1 & \hfil 2 & \hfil 1 & \hfil 10\\
            & ogbn-arxiv-clustered & \hfil 0.003 & \hfil 10,000 & \hfil Tanh & \hfil 2 & \hfil 2 & \hfil 1 & \hfil 10\\
            & reddit-disjoint & \hfil 0.001 & \hfil 10,000 & \hfil Tanh & \hfil 1 & \hfil 2 & \hfil 1 & \hfil 10\\
            \hline
        \end{tabular}
        }
    \label{tab:inductive_results_other_architectures_best_hyperparams}
\end{table}

\clearpage